\documentclass[10pt,twocolumn,letterpaper]{article}

\usepackage{iccv}
\usepackage{times}
\usepackage{epsfig}
\usepackage{graphicx}
\usepackage{amsmath}
\usepackage{amssymb}
\usepackage{multirow}
\usepackage{float}
\usepackage{overpic}
\usepackage{xcolor}
\usepackage{subfigure}
\usepackage{url}            
\usepackage{booktabs}       
\usepackage{amsfonts}       
\usepackage{nicefrac}       
\usepackage{microtype}      
\usepackage{epsfig}
\usepackage{colortbl}
\usepackage{multicol}
\usepackage{multirow}
\usepackage{float}
\usepackage{multirow}
\usepackage{makecell}
\usepackage{cuted}
\usepackage[font={small}]{caption}
\usepackage{pifont}
\usepackage{lipsum} 
\usepackage[misc]{ifsym}

\usepackage[accsupp]{axessibility}

\definecolor{citecolor}{RGB}{65,105,225}
\usepackage[pagebackref=true,breaklinks=true,letterpaper=true,colorlinks,
citecolor=citecolor,bookmarks=false]{hyperref}

\newcommand\blfootnote[1]{%
  \begingroup
  \renewcommand\thefootnote{}\footnote{#1}%
  \addtocounter{footnote}{-1}%
  \endgroup
}

\iccvfinalcopy 

\def\iccvPaperID{3336} 

\begin{document}
\title{OrthoPlanes: A Novel Representation for Better 3D-Awareness of GANs}
\author{
    Honglin He$^{1,2 \dagger ^*}$
    \quad
    Zhuoqian Yang$^{1,3 \ddagger ^*}$
    \quad
    Shikai Li$^{1}$
    \quad
    Bo Dai$^{1}$
    \quad
    Wayne Wu\textsuperscript{1 \Letter}
    \quad \\
    $^{1}$ Shanghai AI Laboratory
    \quad
    $^{2}$ Tsinghua University
    \quad \\
    $^{3}$ School of Computer and Communication Sciences, EPFL
    \\
    {\tt\small hehl21@mails.tsinghua.edu.cn }
    \quad
    {\tt\small zhuoqian.yang@epfl.ch }
    \quad
    {\tt\small lishikai@pjlab.org.cn}
    \\
    {\tt\small daibo@pjlab.org.cn}
    \quad
    {\tt\small wuwenyan0503@gmail.com}
}

\twocolumn[{%
\renewcommand\twocolumn[1][]{#1}%
\maketitle
\begin{center}
    \centering
    \vspace{-7mm}
    \includegraphics[width=1\textwidth,height=58mm]{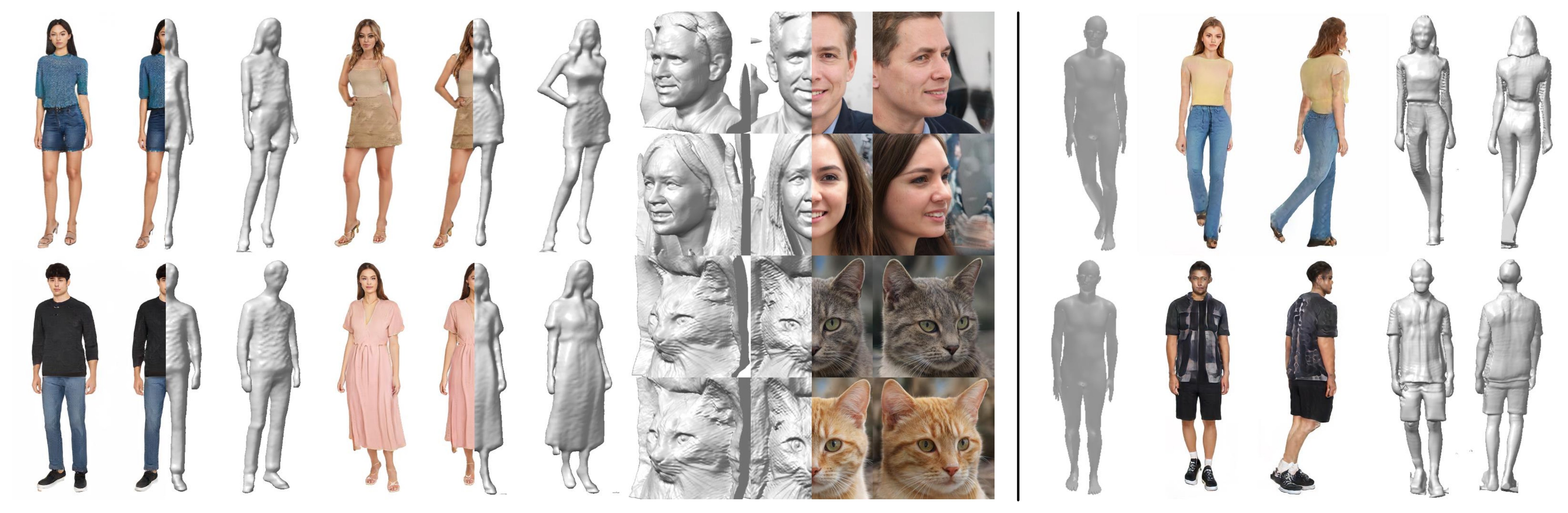}
    \vspace{-3mm}
    \captionof{figure}{\textbf{We build a new hybrid explicit-implicit 3D representation - OrthoPlanes}. Our model can synthesize diverse objects with reasonable geometry without target-specific 3D priors (Left). Furthermore, as a general representation, our approach can be translated to conditional tasks expediently (Right).}
    \vspace{-1mm}
\label{fig:teaser}
\end{center}%
}]

\blfootnote{$^*$ Equal contribution.}
\blfootnote{$\dagger$ Work done as an intern at Shanghai AI Lab.}
\blfootnote{$\ddagger$ Work done as a research engineer at Shanghai AI Lab.}

\begin{abstract}
     We present a new method for generating realistic and view-consistent images with fine geometry from 2D image collections. Our method proposes a hybrid explicit-implicit representation called \textbf{OrthoPlanes}, which encodes fine-grained 3D information in feature maps that can be efficiently generated by modifying 2D StyleGANs. Compared to previous representations, our method has better scalability and expressiveness with clear and explicit information. As a result, our method can handle more challenging view-angles and synthesize articulated objects with high spatial degree of freedom. Experiments demonstrate that our method achieves state-of-the-art results on FFHQ and SHHQ datasets, both quantitatively and qualitatively. Project page: \url{https://orthoplanes.github.io/}.
\end{abstract}

\section{Introduction}

Recovering 3D world from 2D images, known as inverse rendering, is a typical problem in computer vision and computer graphics. It has many practical uses in VR/AR and other domains like movie production and virtual try-on. 

Physically-based methods \cite{collet2015high, dou2016fusion4d} can yield superior outcomes, but they are associated with notable costs and intensive labor requirements. On the other hand, data-centric approaches stand out for their adaptability, ease of use, and photorealism \cite{tewari2020state}. In this study, our objective is to develop a novel data-driven representation that enhances the 2D GAN's understanding of 3D geometry, enabling it to produce more detailed and realistic images of diverse objects.

3D-aware GANs have advanced rapidly to synthesize images that are consistent across multiple views from 2D image collections \cite{chan2021pi, chan2022efficient, gu2021stylenerf, or2022stylesdf, niemeyer2021giraffe, schwarz2020graf, zhang2022avatargen, hong2022eva3d, zhou2021cips, shi2023learning}. These methods combine unsupervised learning, neural rendering and super resolution to produce realistic results. However,  they continue to face challenges in accurately reconstructing 3D shapes from 2D images, a crucial step in rendering 2D GAN genuinely aware of 3D contexts.
Certain methods attempt to address this issue utilizing a tri-plane representation \cite{chan2022efficient} or through the use of multiple parallel images \cite{zhao2022generative} for the reconstruction of 3D entities. However, these strategies fall short when dealing with non-rigid entities like human figures that exhibit asymmetrical poses and varied appearances, which cannot be adequately represented by merely three orthogonal or parallel projections.

\begin{figure}[t]
\centering
\setlength{\abovecaptionskip}{0pt} 
\setlength{\belowcaptionskip}{0pt}
   \includegraphics[width=1.0\linewidth]{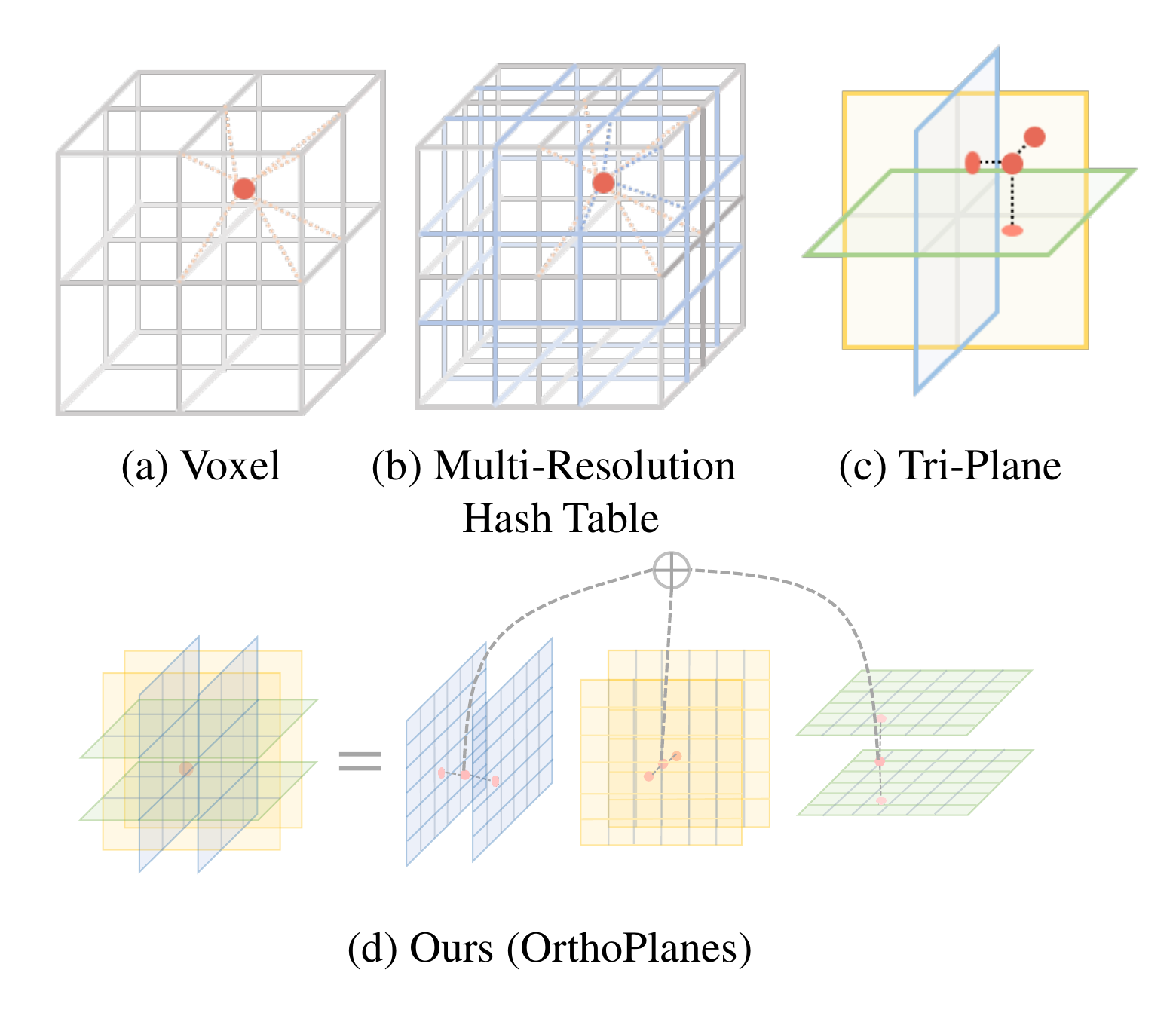}
   \caption{\textbf{Look-up based 3D representations.} Dense voxel grid \cite{sitzmann2019deepvoxels} (a) uses explicitly located anchor-point table to store the scene. Multi-resolution hash table \cite{muller2022instant} (b) uses multiple hash tables to store the at with different resolutions. Tri-plane \cite{chan2022efficient} (c) projects 3D spatial points onto three orthogonal planes, each of which stores compressed features of scene. Our orthoplanes (d) uses three orthogonal plane-groups to store the scene, each plane stores compressed features of the scene but preserves information associated with projection distance.}
\label{fig:representation}
\end{figure}

Inspired by successful scene-specific representations \cite{muller2022instant}, we present a method for efficiently upscaling the generalizable tri-plane representation \cite{chan2022efficient}.
The main idea of our work is that a representation indexed via sectional projection can better represent 3D world than one that is indexed via orthogonal or parallel projection, because it has more explicit 3D information such as projection direction and distance. Based on this intuition, we introduce a novel hybrid explicit-implicit 3D representation, \textbf{OrthoPlanes}, as shown in Fig. \ref{fig:representation} (d). 
We introduce a pre-defined location embedding to each feature map \cite{karras2020analyzing} to make them location-aware. The renderings exhibit better view consistency owing to this explicit data. Moreover, the geometry can be more detailed since the codebook size and density surpasses that of the representations based on orthogonal projections.

With acceptable increase of computing costs over EG3D \cite{chan2022efficient}, our approach can improve image quality, deal with more difficult viewing perspectives and generate articulated objects with high spatial variability, \eg, human bodies, as shown in Fig. \ref{fig:teaser}. Our experiment show state-of-the-art results for 3D-aware image synthesis from 2D collections on diverse datasets including FFHQ \cite{karras2019style} and  SHHQ \cite{fu2022stylegan}. 

We summarize our contributions as the follows: 1) We present a 3D representation termed \textit{orthoplanes} aimed at enhancing the 3D awareness of 2D GANs, which significantly improves view-consistency and geometry.
2) we add a new branch to the 2D generator ensuring both training efficiency and heightened scalability.

\section{Related Works}

\noindent\textbf{Neural scene representation.}  
 Neural scenes optimized through differentiable neural representations from collections of 2D-images and corresponding camera pose \cite{mildenhall2020nerf,barron2022mip,hedman2021baking,tancik2022block,reiser2021kilonerf,fridovich2022plenoxels,lindell2021autoint,liu2020neural,kellnhofer2021neural,schwarz2022voxgraf,jiang2020sdfdiff,zhang2020nerf++,yariv2021volume} and rendered via volume rendering \cite{max1995optical} has been gaining attention. 
The essence of these scene representations revolves around associating each spacial location with a feature vector, which can range from color, density, to signed distance \cite{park2019deepsdf}, among others. Generally, this mapping technique, or the 3D-scene representation, can be categorized into two primary types: point-wise processing \cite{mildenhall2020nerf,zhang2020nerf++,park2019deepsdf} and methods based on lookup procedures \cite{fridovich2022plenoxels,chan2022efficient,muller2022instant}.

Point-wise methods include fully-implicit and local-implicit representations. The fully-implicit representations utilize a coordinate neural network and depict the scene as a continuous function. Although they tend to be efficient in memory usage, rendering a scene using these implicit approaches can be time-consuming in practice. This is because every queried point must be processed through deep neural networks. On the other hand, local-implicit methods \cite{reiser2021kilonerf, jiang2020local, chen2021learning, chabra2020deep} rely on manually designed partitions. A point within a specific partition can query its corresponding shallow neural network to ascertain its features, which offers numerous advantages.

Look-up based methods include explicit and hybrid representations. Explicit representations such as voxel grids \cite{sitzmann2019deepvoxels}, are fast to query but memory unfriendly. Hybrid explicit-implicit representations \cite{muller2022instant, chan2022efficient, martel2021acorn} which bring together the advantages of both forms, are often favored. Conceptually, these approaches serve as codebooks storing the spatial or temporal features \cite{xu2022pv3d, fridovich2023k} of a scene. The representational capacity of these methods is intrinsically tied to the size of their codebooks and spatial distribution of the codes. As illustrated in Fig. \ref{fig:representation}, our proposed representation employs three orthogonal groups of parallel planes to encode the scene. On the one hand, the increased number of planes result in increased density of 3D information. On the other hand, the dispersion of this 3D data facilitates the spread of gradients throughout the 3D scene during training.
\begin{figure*}[htbp]
    \centering
    \includegraphics[width=1.0\textwidth,height=48mm]{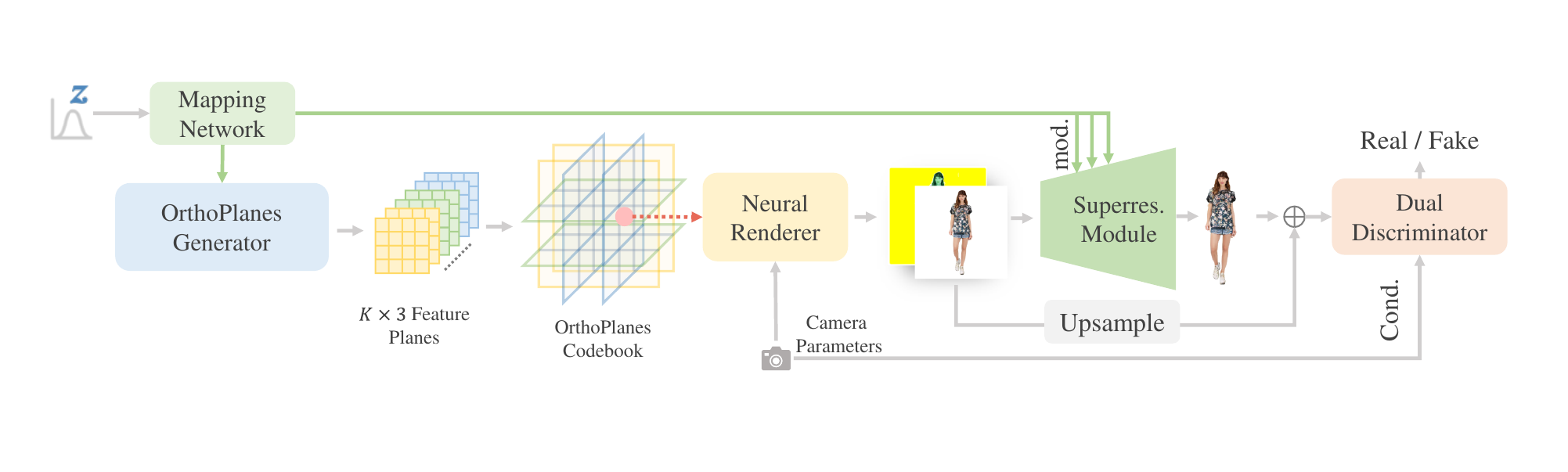}
    \captionof{figure}{\textbf{Our unconditional GAN framework.} Our generator comprises of: a codebook generator with a mapping network, a neural renderer implemented by {\bf orhtoplanes} representation with a lightweight MLP-based feature decoder, a volumetric renderer and a super resolution module. The discriminator employs dual-discriminator design to ensure view-inconsistency. }
\label{fig:pipe}
\end{figure*}

\noindent\textbf{3D-aware image generation.} Generative adversarial networks \cite{goodfellow2020generative} have achieved photo-realistic 2D image quality \cite{karras2019style, karras2020analyzing, karras2021alias}. A recent trend in image synthesis is to generate high-quality multi-view-consistent images with high-fidelity geometry from 2D image collections. Previous works used voxel-based GANs, in which 2D-CNN are replaced by 3D-CNN \cite{gadelha20173d, henzler2019escaping, nguyen2019hologan, nguyen2020blockgan}. Due to the use of voxel grids and 3D convolutions, the memory requirements of these approaches are so high that it makes high-resolution training infeasible. These limitations seen a lift thanks to the development of neural implicit representations \cite{mildenhall2020nerf, schwarz2020graf, chan2021pi}. Nevertheless, since these methods process each spatial points by a deep MLP, it is still challenging to generate images at a resolution achieved by state-of-the-art 2D methods.

For the creation of an effective 3D GAN capable of producing high-resolution images with precise geometry, EG3D \cite{chan2022efficient} was introduced. Leveraging StyleGAN2 as the foundation to generate 2D feature maps that serve as codebooks for point features, and combined with upsampling layers and a dual-discrimination approach, it proves both efficient and expressive. While it successfully delivers high-quality images with detailed geometry for rigid objects, the mere three projections fall short in adequately representing articulated objects possessing a greater spatial complexity.
We discover that the quality of generated image and geometry is associated with the size of the codebooks, not only the resolution but also the spatial distribution features. We are thus motivated to improve the expressiveness of this hybrid explicit-implicit representation by increasing the number of feature planes and placing them evenly across the 3D scene.

Approaches for 3D-aware image synthesis like GMPI \cite{zhao2022generative} and EpiGRAF \cite{skorokhodov2022epigraf} demonstrate impressive image quality. The core difference between these works and ours is that we design a novel 3D representation used for volumetric rendering of generator motivated by enlarging the size of neural codebook while GMPI is based on multiplane image rendering which is 2.5D and EpiGRAF paid attention to developing efficient discriminators for training to synthesize high-resolution images directly based on existing representations.
\section{Methodology}
Our goal is to use orthoplanes in GAN to synthesize high-quality multi-view-consistent image with high-fidelity geometry on diverse datasets. In this section, we first introduce our motivation and the design of orthoplanes  (Sec. \ref{Voxplane}). While improving the expressiveness based on well-defined planes, we maintain the efficiency based on our new design of generator (Sec. \ref{codebook gen}). We discuss inference of our model on both unconditional (Sec. \ref{uncond}) and conditional (Sec. \ref{cond}) generation tasks.

\subsection{OrthoPlanes 3D Representation}
\label{Voxplane}

As discussed in EG3D \cite{chan2022efficient}, to train a high-resolution 3D-aware GAN, the representation should be both efficient and expressive. The tri-plane representation projects each 3D coordinate to three axis-aligned orthogonal feature planes, each with size $N\times N\times C$ where $N$ is the resolution and $C$ is the number of channels. The feature associated with the location $x\in R^3$ is given by $F_{xy}+F_{yz}+F_{zx}$, each summed term extracted from its corresponding plane via bilinear interpolation. The color and density used for volume rendering \cite{mildenhall2020nerf} is obtained by processing this feature with a lightweight MLP. 

From our perspective, a tri-plane is essentially a codebook for storing 3D information efficiently like other methods \cite{sitzmann2019deepvoxels, muller2022instant}. 
For a 3D space with a resolution of $N$, it scales at $O(N^2)$ complexity, which is more efficient than the $O(N^3)$ complexity of voxel \cite{sitzmann2019deepvoxels}.  However, if one wish to enlarge the size of the codebook from $N^2$ to $KN^2$, $K$ must be the power of $4$, and if $K=4^i$, the backbone of StyleGAN2 must be with $i$ more layers, which is inefficient based on the design of StyleGAN2.

In our orthoplanes formulation, we use three axis-aligned orthogonal groups of feature planes, each with $K$ parallel planes with resolution of $N\times N\times C$, as shown in Fig. \ref{fig:representation}d. Each plane is anchored along its projection direction, and distributed evenly along the axis.

We associate each spatial location with a combination of three sectional projections onto the neareast planes on three orthogonal directions. Take $\boldsymbol{F}_{X_fY_fZ_c}$ as an example, it denotes the feature extracted from plane group $X_fY_f$, projected along $Z_c$, using trilineaer interpolation. 

\begin{figure}

\begin{center}
\includegraphics[width=1.0\linewidth]{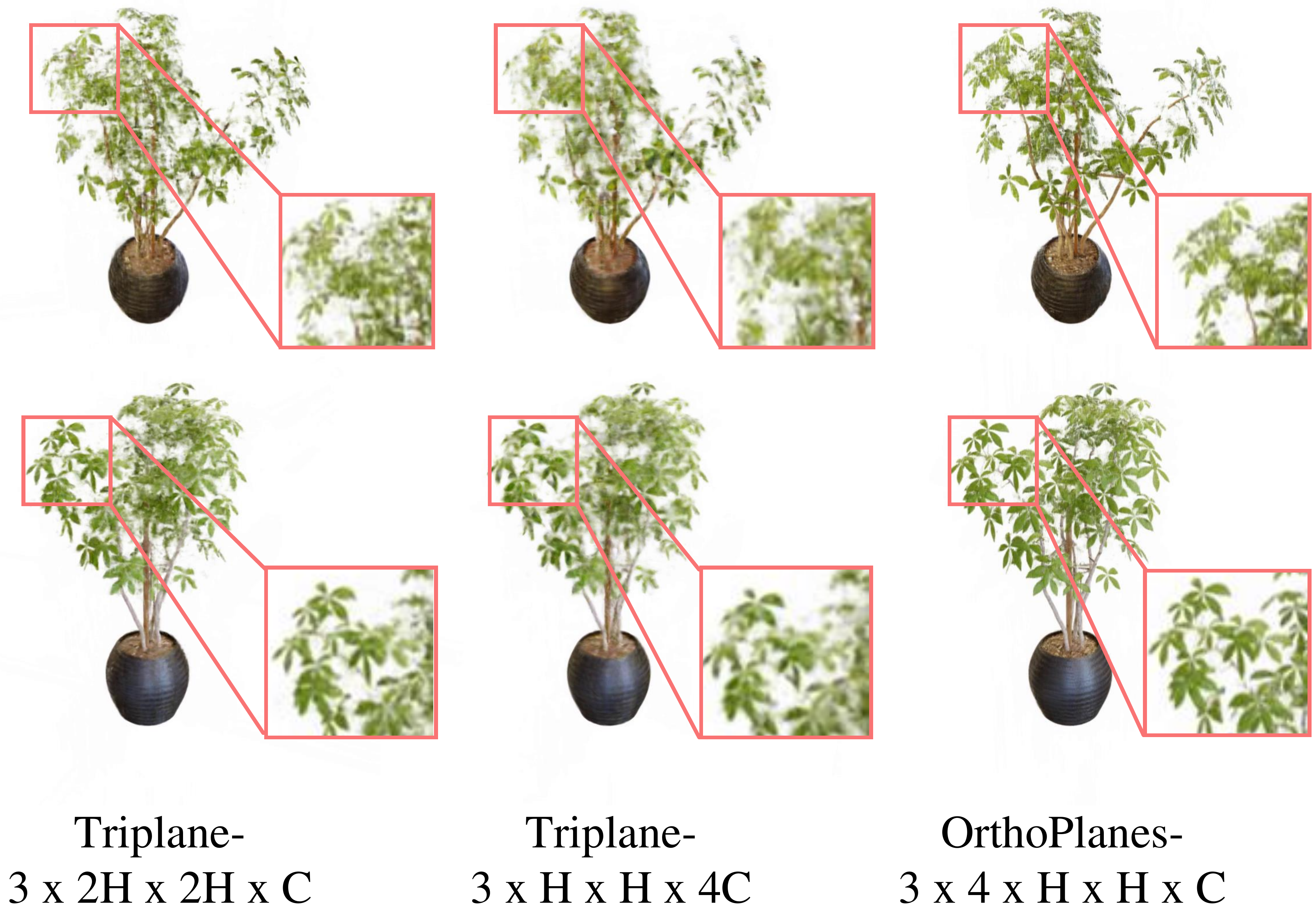}
\end{center}
\vspace*{-4mm}
\caption{\textbf{Single-scene overfitting on the \emph{ficus} scene.} Please zoom in for details.}
\label{fig:nerfficus}
\end{figure}

We demonstrate the expressiveness of orthoplanes with a single-scene overfitting (SSO) experiment. We show quantitative results in Tab. \ref{overfitting} and one example scene in Fig. \ref{fig:nerfficus}. As shown in Tab. \ref{overfitting}, our model archives better PSNR in all cases compared to tri-plane, while maintaining consistent parameter values among methods. As shown in Fig. \ref{fig:nerfficus}, our approach recovers the color tone of \emph{ficus}, especially asymmetric parts like sparse leaves and protruding branches.

Note that our method scales at $O(KN^2)$ complexity, which means we can control the codebook size by manipulating $K$. Meanwhile, just like a voxel with inhomogeneous resolution, it's still much more efficient than voxel grids when $K$ is much smaller than $N$. Furthermore, based on the StyleGAN2 backbone, we can generate these planes efficiently using the method detailed in Sec. \ref{codebook gen}.

\subsection{Codebook Generator}
\label{codebook gen}

The feature codebook of our orthoplanes representation is generated with a StyleGAN2 \cite{karras2020analyzing} generator. The random latent code $\boldsymbol{z}$ sampled from normal distribution is processed by the mapping network to get the style vector $\boldsymbol{w}$, used to modulate the convolution kernels in synthesis layers.

\begin{table}
\begin{center}

\resizebox{1.0\linewidth}{!}{
\begin{tabular}{c c  c  c  c }
\hline
& \emph{lego} & \emph{ficus} & \emph{family} &\emph{caterpillar}\\
\hline

Tri-plane ($H^2,4C$)   &26.56 &23.13 &30.59 &22.03\\
Tri-plane ($(2H)^2,C$)   &28.66 &23.50 &30.48 &22.32\\
Ours ($H^2,C;K=4$) &$\mathbf{31.12}$ &$\mathbf{26.83}$ &$\mathbf{32.06}$ &$\mathbf{23.93}$\\
\hline
\end{tabular}}
\end{center}
\vspace*{-4mm}
\caption{\textbf{Quantitative evaluation of PSNR on SSO cases.}}
\vspace*{-6mm}
\label{overfitting}
\end{table}

To make the plane location-aware, we introduce a location embedding $\boldsymbol{l}_i$ to mark the position of each plane. 
Each plane-group is a set of tuples $(\boldsymbol{I}_i, \boldsymbol{l}_i),i\in \left\{1,2,...,K\right\}$, where $\boldsymbol{I}_i\in R^{H\times W\times C}$ denotes the feature map of the $i$-th plane, while $\boldsymbol{l}_i \in [-1,1]$ denotes the location embeddings of the $i$-th plane along corresponding projection direction.
In practice, we use a  StyleGAN2  generator to generate tensors of shape $K\times N\times N\times 3C$, and reshape it to $3\times K\times N\times N\times C$ by split and permutation to obtain the three orthogonal plane groups, where $N$ is the resolution and $3C$ the number of channels.

\begin{figure}[t]
\centering
\setlength{\abovecaptionskip}{0pt} 
\setlength{\belowcaptionskip}{0pt}
\includegraphics[width=0.85\linewidth]{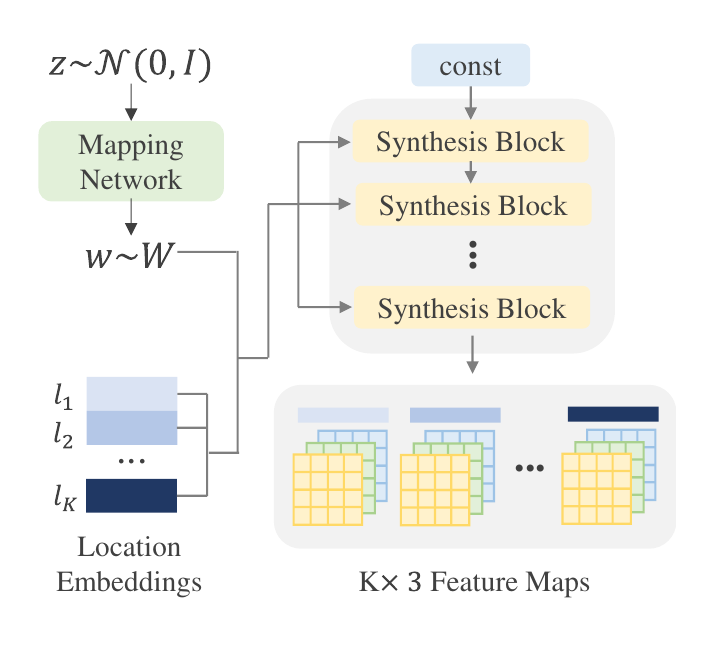}
\caption{\textbf{Architecture of orthoplanes generator.} The structure of synthesis blocks follow StyleGAN2 \cite{karras2020analyzing}. We replace the ToRGB layer by our ToFeature layer to generate location-embedded feature maps in parallel.}
\label{fig:block}
\end{figure}

\begin{figure}[t]
\centering
\includegraphics[width=0.85\linewidth]{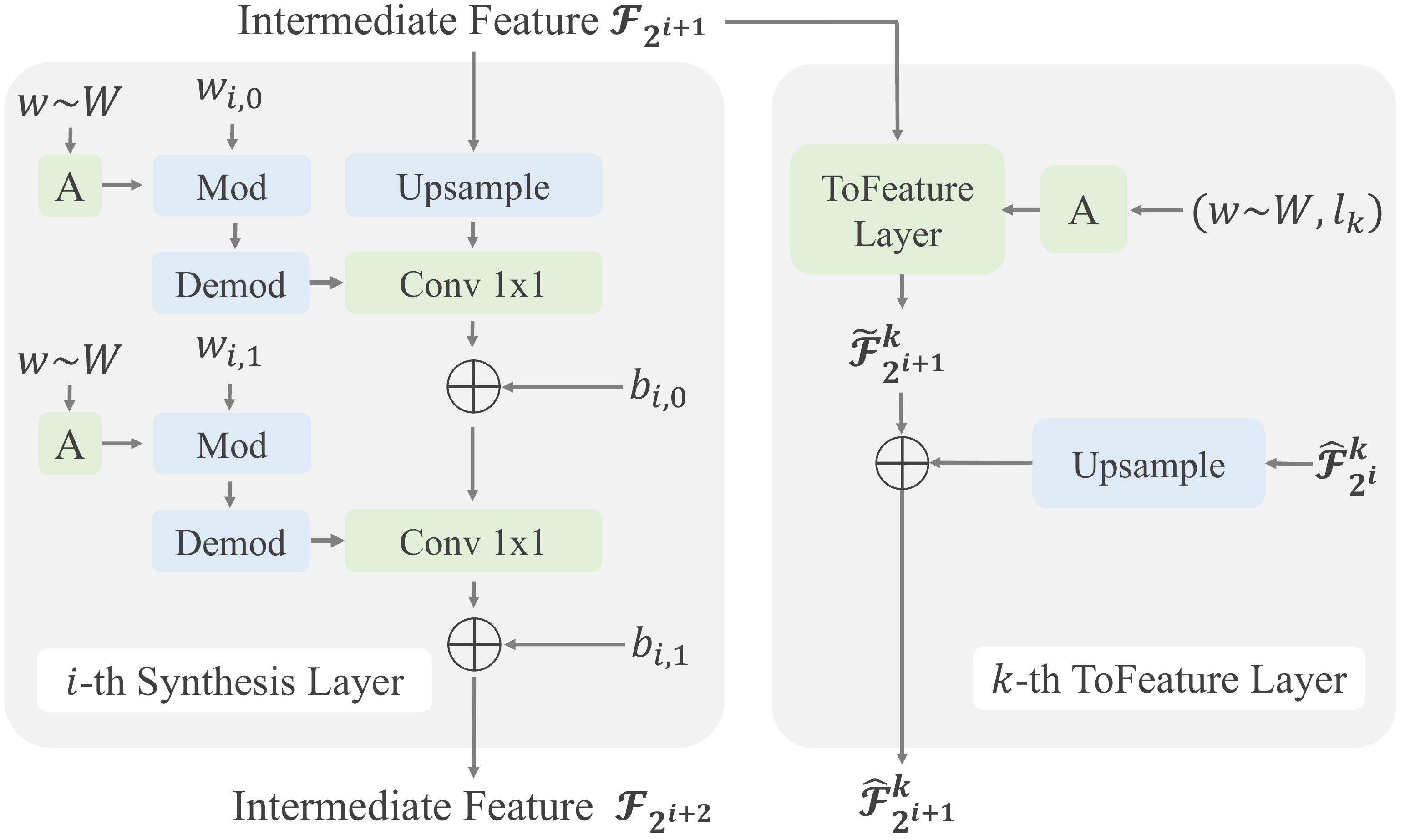}
\caption{\textbf{Implementation of our generator.} There are $K$ parallel ToFeature layers with shared weights in a single synthesis layer. We reuse the intermediate feature map synthesized by StyleGAN2 synthesis layer to ensure efficiency.}
\vspace*{-4mm}
\label{fig:synlayer}
\end{figure}

The same process can be viewed as generating $K$ tri-planes in parallel. Inspired by GMPI \cite{zhao2022generative}, we choose to reuse the same intermediate feature map synthesized in each synthesis block across all tri-planes, repeat them $K$ times, and modulate the results in \textit{ToRGBLayer} using the style vector $\boldsymbol{w}$ with location embeddings $\boldsymbol{l}_i$. Specifically, we use $\mathcal{R}=\left\{4,8,16,...,N\right\}$ to denote the resolution of the results synthesized by each synthsis block. The intermediate feature maps at each resolution $h\in\mathcal{R}$ are computed with a synthesis block $f_{Syn,h}$, \ie, 

\begin{equation}
    \label{stylegan inter map}
    \mathcal{F}_h=f_{Syn,h}(\boldsymbol{w}), \forall h\in \mathcal{R}
\end{equation}

\begin{figure*}[htbp]
    \centering
    \includegraphics[width=0.95\textwidth,height=78mm]{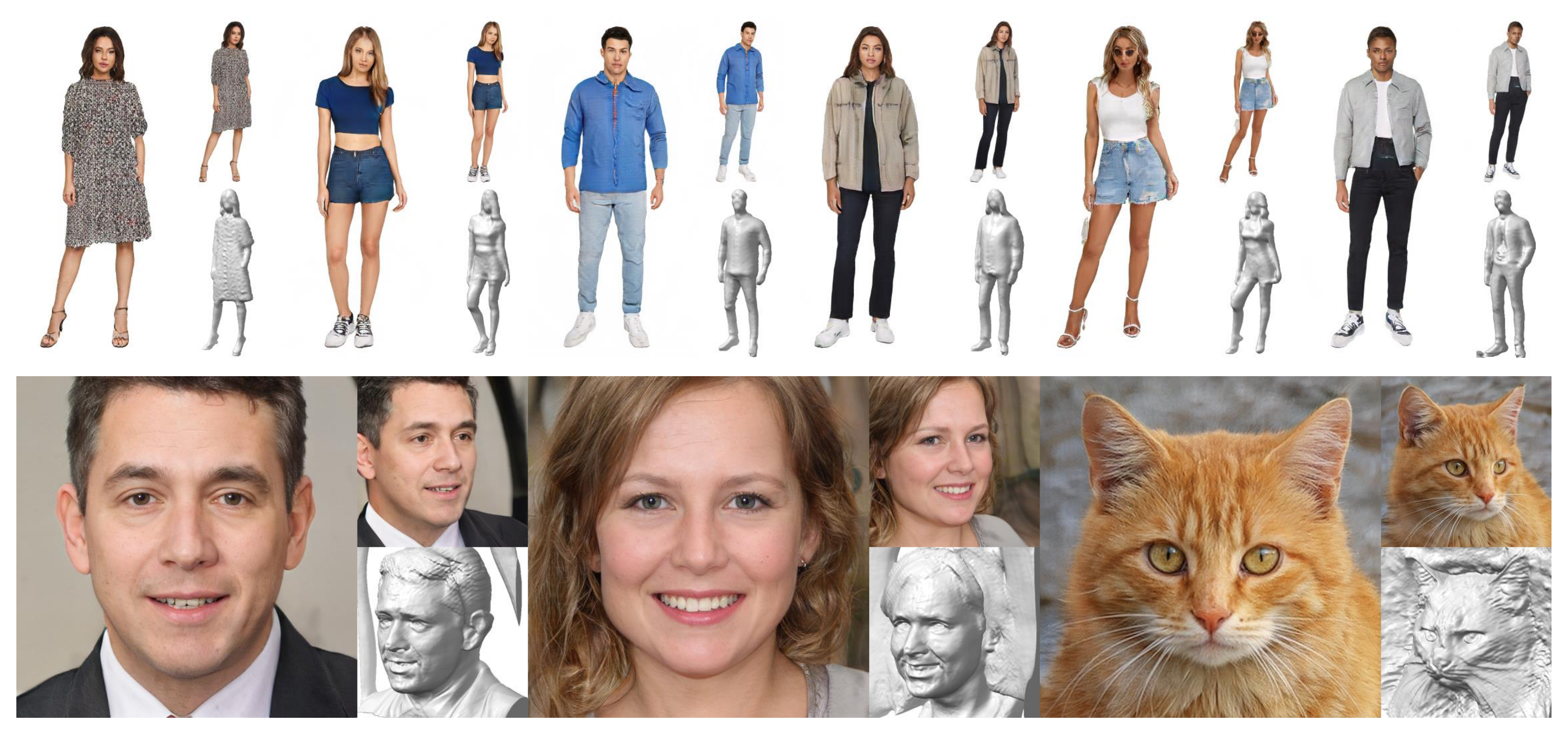}
    \captionof{figure}{\textbf{Curated examples at $512^2$.} (we clip images of human bodies to the resolution of $512\times256$), synthesized by models trained with SHHQ \cite{fu2022stylegan}, FFHQ \cite{karras2019style} and AFHQv2-Cat \cite{choi2020stargan}.}
\label{fig:qual}
\end{figure*}

The $i$-th feature codebook $\hat{\mathcal{F}}^i_N\in R^{N\times N\times 3C}$ used in volumetric rendering, with resolution of $N$, is obtained by accumulating intermediate results at all resolutions, \ie, 

\begin{equation}
    \label{stylegan codebook map}
    \hat{\mathcal{F}}^i_h= \left \{
    \begin{array}{ll}
    \widetilde{\mathcal{F}}^i_h + Upsample(\hat{\mathcal{F}}^i_{\frac{h}{2}}),                    & h\in\mathcal{R} \backslash \left\{4\right\} \\
    \widetilde{\mathcal{F}}^i_h,     & h=4
    \end{array}
\right.
\end{equation}
where $\widetilde{\mathcal{F}}^i_h$ is the residual given by \textit{ToFeature} Layer $f_{ToFeat,h}$ at resolution $h$, and $\widetilde{\mathcal{F}}^i_h$ the feature codebook. $f_{ToFeat,h}$ takes $\mathcal{F}_h$ and concatenated vector $(\boldsymbol{w},\boldsymbol{l}_i)$ as input, to get the location-specific feature codebook. Based on this design, $f_{ToFeat,h}$ operates on each tri-plane individually, and the operation can be done in parallel. Furthermore, it uses the same intermediate feature map $\mathcal{F}_h$ as input. These characteristics preserve the efficiency while expanding the expressive power of 3D representation. In Table. \ref{shhq-100k-new}, compared to enlarging the resolution or the number of channels of the generated feature map, our method can generate larger codebooks with less memory and time. 

\begin{table}

\begin{center}
\resizebox{\linewidth}{!}{
\begin{tabular}{c|  c c c c}
\hline
 & FID $\downarrow$ & \# FLOPs $\downarrow$ & T.Speed $\uparrow$ & T.Mem. $\downarrow$\\
\hline
EG3D ($256^2, 32$)   &$14.1$  &126.25G& ${1\times}$  &${1\times}$ \\
EG3D ($512^2, 32$)  &$14.2$ ($+0.7\%$)  &150.14G($+18.9 \%$) &$0.51\times$ &$1.55\times$  \\
EG3D ($256^2, 128$)  &11.6 ($-17.7\%$)&149.38G($+18.3 \%$) &$0.41\times$ &$1.75\times$   \\
Ours ($256^2, 32;K=4$)  &$\mathbf{11.2}$ ($-20.6\%$) &129.57G($+2.6 \%$)&${0.82\times}$ &${1.16\times}$   \\
\hline
\end{tabular}}
\end{center}
\vspace*{-4mm}
\caption{\textbf{Quantitative evaluation on 100K-SHHQ.} \textit{CB. size} denotes relative codebook size, i.e. $K \times H^2 \times C$. \textit{T.Speed} and \textit{T.Mem} denote the speed and GPU memory usage during training, respectively. \textit{I.Speed} denotes inference speed. \textit{CB.Size}, \textit{T.Speed}, \textit{T.Mem} and \textit{I.Speed} are all reported relative to the EG3D ($256^2,32$) model.}
\vspace*{-4mm}
\label{shhq-100k-new}
\end{table}

\subsection{Unconditional GAN Framework}
\label{uncond}

Fig. \ref{fig:pipe} gives an overview of our generator architecture for unconditional 3D-aware image synthesis,. Orthoplanes representation is used as the codebook for spatial query during volumetric rendering. 
Following \cite{niemeyer2021giraffe, chan2022efficient}, the output of volumetric renderer is a feature image $\boldsymbol{I}_F$ and a RGB image $\boldsymbol{I}_{RGB}^-$, both in low resolution. A super resolution module upsamples these images to yeild the final high-resolution RGB image $\boldsymbol{I}_{RGB}$.

\noindent{\bf Orthoplanes Generator.} The codebook for orthoplanes is produced using a StyleGAN2 generator, as outlined in \cite{karras2020analyzing}, incorporating our supplementary branch detailed in Sec. \ref{codebook gen}. Initially, the random latent code $\boldsymbol{z}$ is sampled from a standard normal distribution $\mathcal{N}(0,I)$. This is subsequently transformed by a mapping network to obtain the intermediate latent code $\boldsymbol{w}$ from the distribution $\mathcal{W}$. This code then modulates the weights of the convolution kernels within the codebook synthesis network.

The output of the synthesis network is of shape of $K\times 256\times 256\times 96$. We obtain orthoplanes with shape $3\times K\times 256\times 256\times 32$ through channel-wise split and permutation.

\noindent{\bf Feature Aggregation.} To render an image of the 3D scene, we sample spatial locations olong camera rays calculated with camera parameters $\boldsymbol{P}$, and query their features from the orthoplanes codebook.

\begin{figure*}[htbp]
    \centering
    \includegraphics[width=0.95\textwidth,height=50mm]{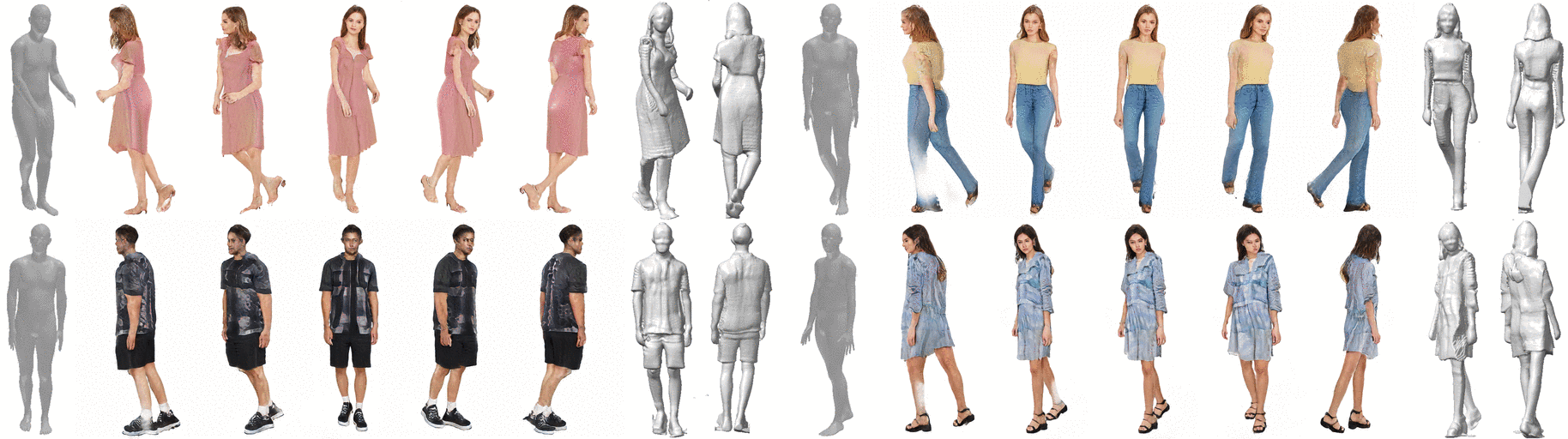}
    \captionof{figure}{\textbf{Curated examples on conditional SHHQ \cite{fu2022stylegan}.} (we clip images to the resolution of $512\times256$), synthesized by models trained with SHHQ \cite{fu2022stylegan}. Meshes on the left side are visualization of SMPL \cite{loper2015smpl}.}
\label{fig:qual-cond-shhq}
\end{figure*}

The feature of a spatial location from each plane-group is obtained through trilinear interpolation. The features from three orthogonol plane-groups are summed to yield the aggregated feature. This aggregate feature is then processed using a streamlined MLP decoder. Similar to the approach in EG3D \cite{chan2022efficient}, the MLP input excludes any point-wise data such as positional encoding.

\noindent{\bf Volumetric Rendering and Super Resolution.} 
The volumetric renderer is developed with the importance sampling strategy, as highlighted in \cite{mildenhall2020nerf}. In line with other studies \cite{niemeyer2021giraffe, chan2022efficient}, the renderer yields a low-resolution RGB image, denoted as $\boldsymbol{I}_{RGB}^-$, accompanied by a feature image $\boldsymbol{I}_{F}$. These images are subsequently refined by a super-resolution module to produce the final RGB image, $\boldsymbol{I}_{RGB}$. This super-resolution module comprises of two StyleGAN2 synthesis blocks, which are modulated using the style vector $\boldsymbol{w}$. In all our tests, the input to the super-resolution module has a resolution of $128^2$ and a channel size of $32$.

\noindent{\bf Dual Discrimination.} As introduced in EG3D \cite{chan2022efficient}, we use dual discrimination to enforce view inconsistency, and use the camera parameters $\boldsymbol{P}$ as condition of the discriminator.

Our models in all unconditional experiments use non-saturating GAN Loss \cite{goodfellow2020generative} with R1 regularization \cite{mescheder2018training}.

\begin{align}
     \mathcal{L}_{adv,D} & = \mathbf{E}_{\boldsymbol{z}\sim P_z,  \boldsymbol{c}\sim P_{data}} [f(D(G(\boldsymbol{z};\boldsymbol{c}))) ] \\ & + \mathbf{E}_{\boldsymbol{I}\sim P_{data}} [f(D(\boldsymbol{I})) + \lambda ||\nabla_{\boldsymbol{I}}D(\boldsymbol{I})||^2_2] \\
     \mathcal{L}_{adv,G} &= \mathbf{E}_{\boldsymbol{z}\sim P_z,  \boldsymbol{c}\sim P_{data}} [f(-D(G(\boldsymbol{z};\boldsymbol{c}))) ]
\end{align}

where $f(x)=-\log(1+\exp({-x}))$, $P_{data}$ is the real data distribution and $\boldsymbol{c}$ the camera parameter.

\subsection{Conditional GAN Framework}
\label{cond}

We also examine the effectiveness of our representation for generating articulated objects. Specifically, we build a pose-conditional full-body human image generation model with orthoplanes.

The model architecture of our pose-conditional generator is given in the supplementary materials. The pose parameters $\theta$ defined by SMPL \cite{loper2015smpl} are used as the condition for both of generator and discriminator. We follow the practice of transforming queried coordinates to canonical human-body space \cite{peng2021neural, weng2022humannerf, bergman2022generative, hong2022eva3d, zhang2022avatargen}.

\noindent{\bf Pose-Canonical Human Generation.} We leverage the orthoplanes representation for generation of clothed humans in the canonical space $\mathcal{C}$. Specifically, the random latent code $\boldsymbol{z}$ is used to generate the canonical feature of clothed humans via codebook generator as described in Sec. \ref{codebook gen}. Every point sampled in observation space $\mathcal{O}$ will be transformed to the canonical space $\mathcal{C}$ before querying the codebook.

\noindent{\bf Delta SDF Learning Strategy.} Like concurrent works \cite{hong2022eva3d, zhang2022avatargen}, we use employ the Delta SDF \cite{yifan2021geometry} learning strategy instead of predicting density directly. Specifically, the spatial point $\boldsymbol{x}$ will first compute its signed distance $d_0(\boldsymbol{x})$ from the SMPL Template mesh under given pose $\boldsymbol{\theta}$, instead of output the density $\sigma(\boldsymbol{x})$, we just output the offset $\Delta d(\boldsymbol{x})$ from $d_0(\boldsymbol{x})$ \cite{yifan2021geometry}, and translate SDF value $d_0(\boldsymbol{x}) + \Delta d(\boldsymbol{x})$ to the density value $\sigma(\boldsymbol{x})$ using method proposed by StyleSDF \cite{or2022stylesdf}.

\noindent{\bf Training.} Please see the supplement for implementation details of our conditional GAN framework.

\section{Experiments and Results}
\noindent{\bf Datasets.} To illustrate the effectiveness of our method, we compare with state-of-the-art methods in both unconditional and conditional 3D-aware image generation. For the task of unconditional generation, we compare performance on three datasets FFHQ \cite{karras2019style}, AFHQv2-Cat \cite{choi2020stargan} and SHHQ \cite{fu2022stylegan} separately. FFHQ is a real-world human-face dataset. AFHQv2-Cat is a small cat-face dataset. SHHQ is a real-world human-body dataset with about 220K images.  For FFHQ and AFHQv2-Cat, we use the version with horizontal flips provied by EG3D \cite{chan2022efficient}.  For AFHQv2-Cat, we apply transfer learning \cite{karras2020training} from FFHQ checkpoint  provided by EG3D \cite{chan2022efficient} with adaptive data augmentation \cite{karras2020training}.

\begin{figure*}
    \centering
    \vspace{-2ex}
    \includegraphics[width=0.95\textwidth,height=78mm]{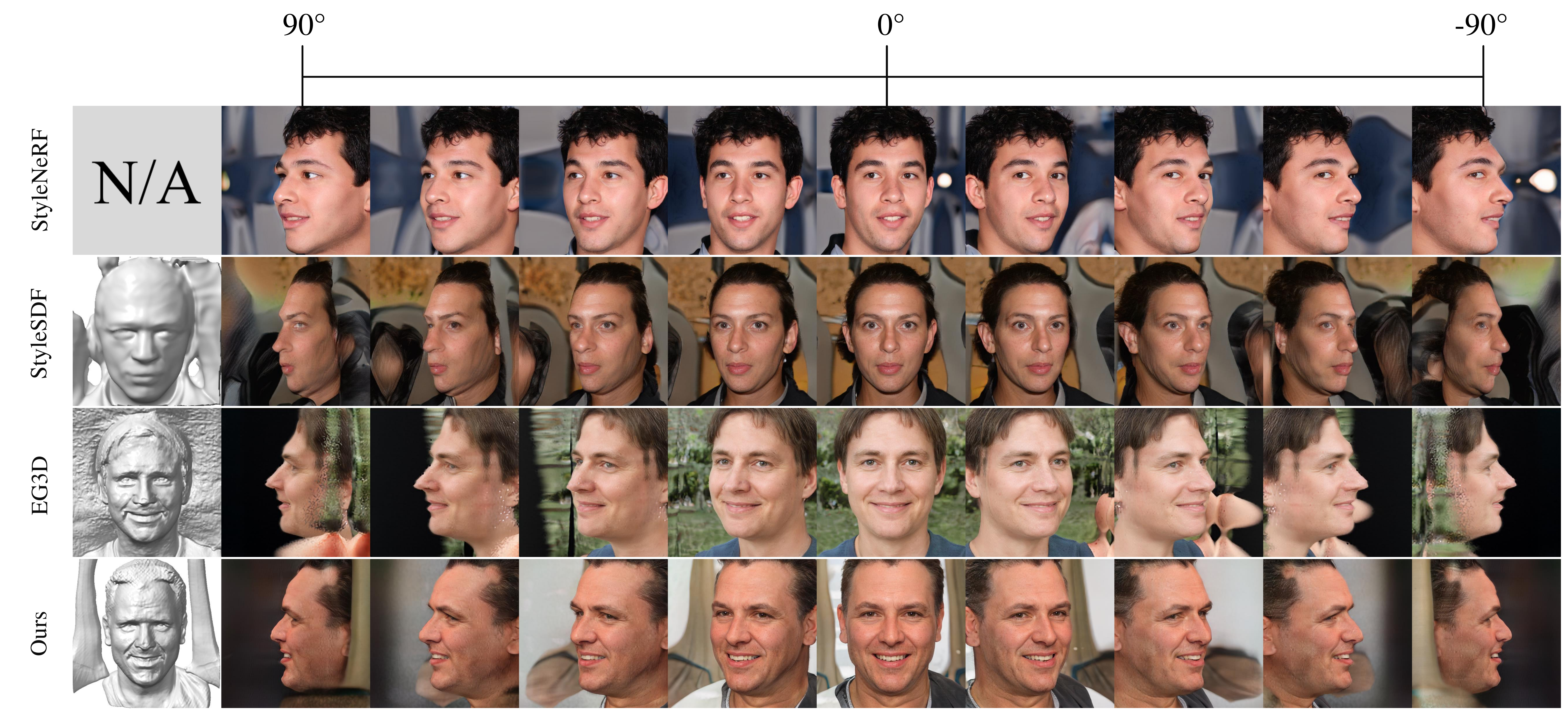}
    \vspace{-1ex}
    \captionof{figure}{\textbf{Images rendered on various camera poses.} There are obvious distortions on  extreme yaw angles in previous works. In contrast, our method is better on both of 2D image quality and 3D geometry.}
    \vspace{-1ex}
    
\label{fig:bigview}
\end{figure*}

For the task of conditional generation, we compare on the full-body human image dataset SHHQ, SMPL pose-condition for each image is estimated using an off-the-shelf model PARE \cite{kocabas2021pare}.  

\noindent{\bf Baselines.} We compare our method against nine state-of-the-art methods for 3D-aware image synthesis.
For unconditional generation, we compare against five methods: $\pi$-GAN \cite{chan2021pi}, StyleSDF \cite{or2022stylesdf}, StyleNeRF \cite{gu2021stylenerf},  GMPI \cite{zhao2022generative} and EG3D \cite{chan2022efficient}.
For conditional GAN, we compare our method against four methods: ENARF-GAN \cite{noguchi2022unsupervised}, 3DHumanGAN \cite{yang20223dhumangan}, EVA3D \cite{hong2022eva3d} and AvatarGen \cite{zhang2022avatargen}.

\noindent{\bf Evaluation Metrics.} We follow prior work to assess the results using common metrics. To measure 2D qualities, we report Fr\'echet Inception Distance \cite{heusel2017gans} (FID) and Kernel Inception Distance \cite{binkowski2018demystifying} (KID). For unconditional FFHQ, we report the  identity consistency (ID) \cite{deng2019arcface, deng2019accurate, chan2022efficient} to measure the view-consistency of generated results. For unconditional SHHQ, we report the depth accuracy (Depth) by calculating Mean Squared Error (MSE) between pseudo depth-maps estimated from generated RGB images by MiDaS \cite{ranftl2020towards} and the generated ones. For conditional SHHQ, we additionally report Percentage of Correct Keypoints (PCK) \cite{noguchi2022unsupervised} to evaluate the effectiveness of the pose controllability. 

\begin{table}
\large
\begin{center}
\resizebox{\linewidth}{!}{
\begin{tabular}{c c c  c | c c}
\hline
& \multicolumn{3}{c|}{FFHQ} & \multicolumn{2}{c}{AFHQv2-Cat} \\
 & FID $\downarrow$ & KID $\downarrow$ &ID $\uparrow$  & FID $\downarrow$ & KID $\downarrow$ \\
\hline
pi-GAN $128^2$  &29.9 &35.73 &0.67  &16.0 &14.92 \\
StyleSDF $256^2$ &11.5 &2.65 &-  &12.8$^*$ &4.47$^*$ \\
StyleNeRF $512^2$ &7.80 &2.20 &-  &13.2$^*$ &3.60$^*$ \\
GMPI $512^2$ &8.29 &4.54 &$\underline{0.74}$  &7.79 &4.74 \\
EG3D $512^2$ &$4.70^\ddagger$ &1.32 &$\boldsymbol{0.77}$ &$\boldsymbol{2.77}^\dagger$ &$\boldsymbol{0.41}^\dagger$ \\
\hline
Ours-S  $512^2$ &$\underline{4.11}^\ddagger$  &$\boldsymbol{1.05}$ &0.71  &$\underline{2.82^\dagger}$ &$\underline{0.46^\dagger}$ \\
Ours-R  $512^2$ &$\boldsymbol{4.01}^\ddagger$ &$\underline{1.23}$ &0.73  &- &-\\
\hline
\end{tabular}}
\end{center}
\vspace*{-6mm}
\caption{\textbf{Quantitative evaluation for FFHQ and AFHQv2-Cats.} 
}
\label{uf}
\end{table}

\blfootnote{For KID, we report all results $\times10^3$.}
\blfootnote{$*$ Metrics are reported on the whole AFHQv2 dataset.}
\blfootnote{$\dagger$ Trained with adaptive discriminator augmentation \cite{karras2020training}.}
\blfootnote{$\ddagger$ Trained with camera condition \cite{chan2022efficient}.}

\subsection{Comparisons}

\begin{figure}[t]
   \centering
   \setlength{\itemsep}{0pt}
\setlength{\parsep}{0pt}
   \includegraphics[width=1.0\linewidth]{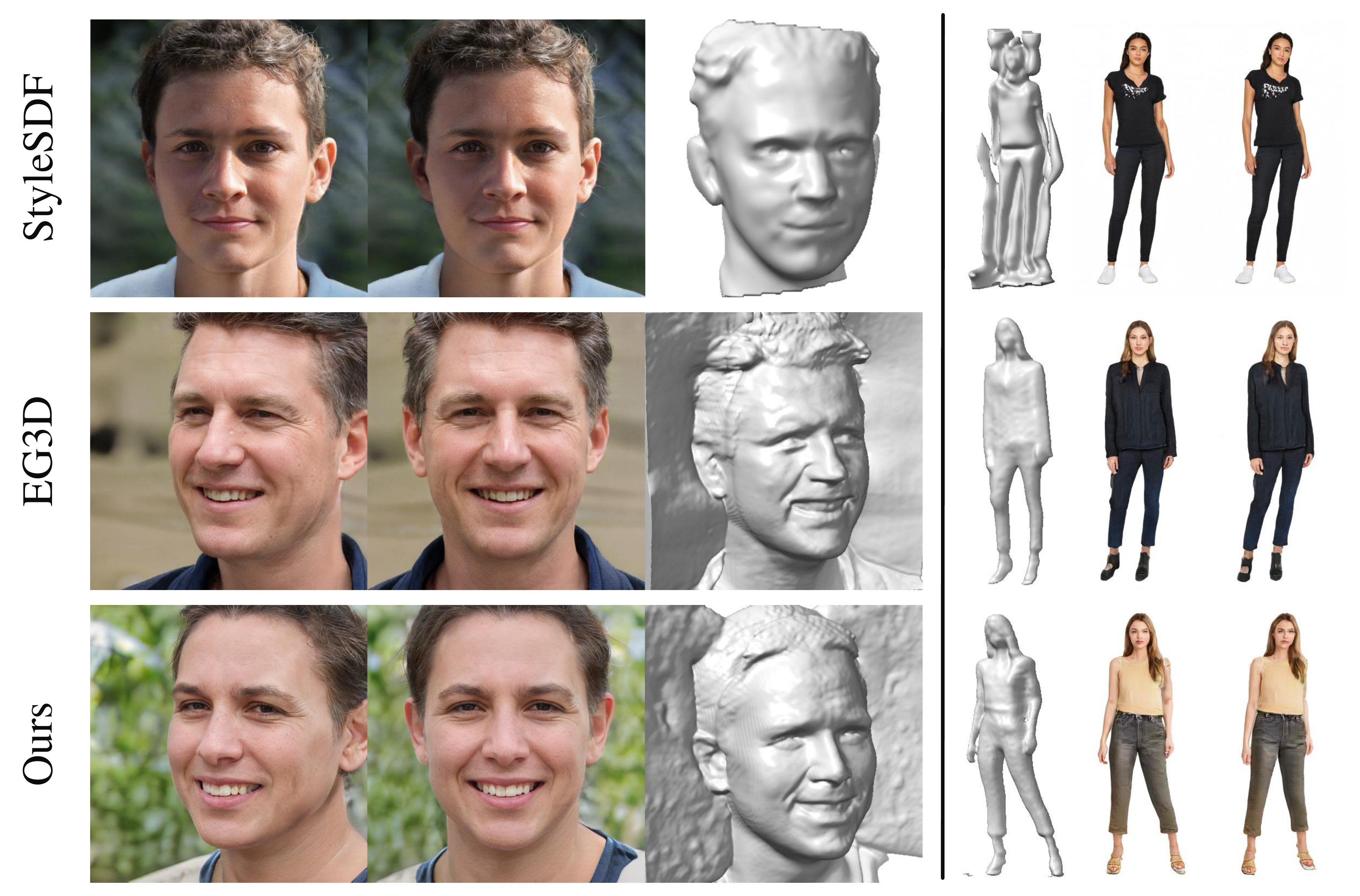}
   \caption{\textbf{Qualitative comparison on FFHQ and SHHQ.} Shapes are extracted by marching cubes. }
\label{fig:compare}
\end{figure}

\noindent{\bf Qualitative Results.} Fig. \ref{fig:qual} presents selected examples synthesized by our model with different datasets. And Fig. \ref{fig:qual-cond-shhq} presents selected examples synthesized by our conditional human body generative model. Fig. \ref{fig:compare} and Fig. \ref{fig:bigview} provide qualitative comparisons against baselines. StyleSDF, EG3D and ours are all based the two-stage approach, which uses low-resolution neural rendering followed by super resolution modules. StyleSDF is a fully-implicit method while EG3D is hybrid explicit-implicit. Geometry generated by SDF is overly smooth and lack of high-frequency details. The geometry of EG3D on SHHQ lack of details because the dataset is extremely unbalanced on viewpose, the majority of data is front view that it will converge to the state that feature only depends on one projection plane. As shown in Fig. \ref{fig:compare}, the geometry of EG3D is too flat while ours is with the feeling of depth thanks to our design of sectional projection.

\begin{table}
\tiny
\begin{center}
\resizebox{\linewidth}{!}{\begin{tabular}{c c c c}
\hline
 & {FID~$\downarrow$}  & {KID~$\downarrow$} & {Depth~$\downarrow$} \\
\hline
{StyleSDF \scalebox{0.7}{${512^2}$}} &{33.29} &{25.2} & \textbf{0.036} \\
{StyleNeRF \scalebox{0.7}{$512^2$}} &{7.60} &{3.96} &-   \\
{EG3D \scalebox{0.7}{$512^2$}}  &{5.79} &{2.26} &{0.082}\\
\hline
{Ours \scalebox{0.7}{$512^2$}} & {\textbf{4.18}} & \textbf{2.05} &{0.082}\\
\hline
\end{tabular}}

\end{center}
\vspace*{-6mm}
\caption{\textbf{Quantitative evaluation for SHHQ.} All models are trained from scratch until convergence.}
\label{ub}
\end{table}

\begin{table}
\large
\begin{center}
\resizebox{\linewidth}{!}{
\begin{tabular}{c c c c c}
\hline
 & FID $\downarrow$ & KID $\downarrow$ & Depth $\downarrow$ &PCK $\uparrow$ \\
\hline
ENARF-GAN $128^2$ &20.09 &16.94 &0.086 &79.40 \\
3DHumanGAN $^\dagger$ $512^2$ &9.31 &$\underline{5.16 }$  &- &- \\
EVA3D $^\dagger$ $512^2$ &11.99 &9.00 &$\boldsymbol{0.017 }$ &88.95 \\
AvatarGen $^\dagger$ $^\ddagger$ $512^2$ &$\boldsymbol{4.29}$ &- &0.365 &$\boldsymbol{99.49 }$ \\
\hline
Ours $512^2$ &$\underline{9.00}$ &$\boldsymbol{4.37}$ &$\underline{0.032 }$ &$\underline{99.12}$  \\
\hline
\end{tabular}}
\end{center}
\vspace*{-6mm}
\caption{\textbf{Quantitative evaluation for conditional SHHQ.}}
\label{cb}
\end{table}

When it comes to view-consistency, as shown in Fig. \ref{fig:bigview}. Our work performs significantly better than previous works. Our work is capable of rendering at a consistent high quality at large yaw angles while previous methods fail, showing different levels of degradation and artifacts. StyleNeRF \cite{gu2021stylenerf} and StyleSDF \cite{or2022stylesdf} can only capture the information within the distribution of dataset. EG3D \cite{chan2022efficient} can synthesize reasonable results within larger range. However, limited by the expressiveness of orthogonal projection and the sparse distribution of features and gradients in space, it fails on extreme yaw angles. With the predefined sectional projection with projection distance, our method captures more spatial information. As an additional benefit, the geometry of human face is separated from the background without additional process, as shown in Fig. \ref{fig:qual} and Fig. \ref{fig:bigview}.

\noindent{\bf Quantitative Results.} Table. \ref{uf} and Table. \ref{ub} provide quantitative metrics comparing the proposed approach and baselines on unconditional generation. For FFHQ and AFHQv2-Cat, Ours-S and Ours-R are models trained from scratch and fine-tuned on pretrained checkpoint provided by EG3D \cite{chan2022efficient}, respectively. In the config-R, the weights of our new branch are randomly initialized, after training about $14\%$ of EG3D's full pipeline training time, it demonstrates notable improvements in FID scores. For SHHQ, our method is the best on FID and KID, for the Depth metrics, StyleSDF is better thanks to higher resolution of neural renderer. To eliminate the possible impact of data volume, we train a model with a small subset of SHHQ with 40K images , results are given in the supplementary materials. For AFHQv2-Cats, since we use the checkpoint of FFHQ from EG3D \cite{chan2022efficient}, there are some key weights of ToFeature layer that cannot be transferred from a larger dataset, limiting its performace. Additionally, Table. \ref{cb} provides quantitative metrics on conditional generation, more results and analysis are given in the supplementary materials. 

\blfootnote{
$^\dagger$ We quota results from their papers.
}

\blfootnote{
$^\ddagger$ AvatarGen uses 50K fake images and less than 50K real images to compute FID, which may cause difference.
}


\subsection{Ablation Study}

\begin{figure}[t]
\begin{center}
   \includegraphics[width=0.95\linewidth]{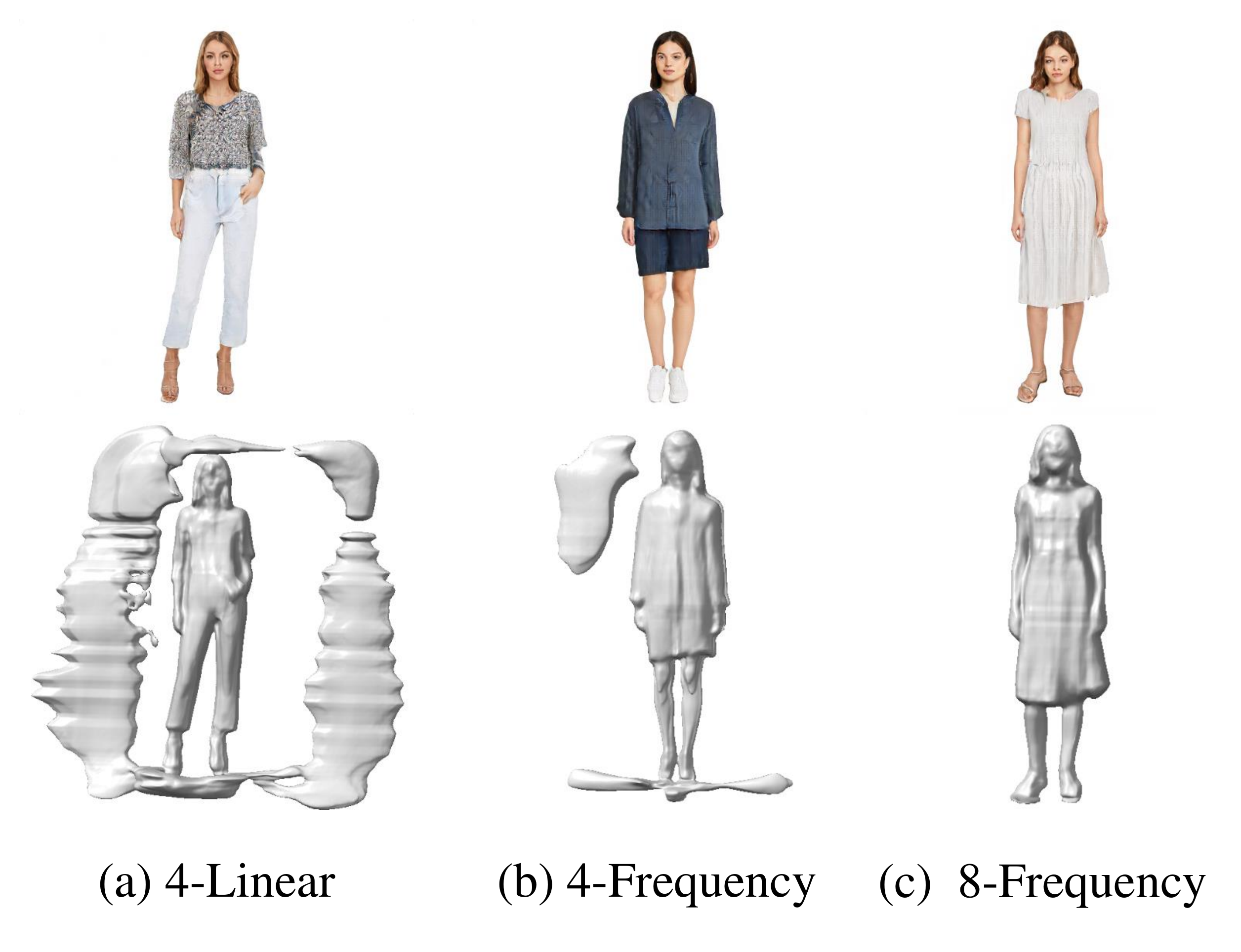}
\end{center}
   \caption{\textbf{Ablation study.} With frequency encoding, there will be less artifacts. By increasing the number of planes, there will be more details and none artifacts on geometry.}
\label{fig:ab}
\end{figure}

\noindent{\bf Location embeddings.} As described in Sec. \ref{codebook gen}, we attach a location embedding to each style vector in the orthoplane codebook. Using a linear embedding $\boldsymbol{l}_k = -1 + \frac{k - 1}{K}\cdot 2$ 
is able to distinguish these planes. If we want to learn finer geometry details, high-frequency information between these planes is needed, for which we use $FE(\boldsymbol{l}_k)$ as location embeddings in majority of our experiments. Intuitively, frequency encoding is equivalent to assigning higher weights for the location embeddings. It's a simple but useful solution to represent the difference among images with different projection distance. As shown in Table. \ref{ab1} and Fig. \ref{fig:ab}, frequency encoding provides improvement in FID and geometry.

\noindent{\bf Number of planes.} Intuitively, network with more planes can capture more information. What's more, based on the flexible design of ToFeature branch, the number of planes can be scaled up easily even after training. As shown in Table. \ref{ab2}, using checkpoint provided by EG3D \cite{chan2022efficient} (1-plane) with some minor additional training, we can obtain better results. As shown in Table. \ref{ab1} and Fig. \ref{fig:ab}, the increase of planes brings improvement on FID and geometry. 

\begin{table}
\normalsize
\begin{center}
\resizebox{\linewidth}{!}{
\begin{tabular}{c c c c c c}
\hline
  &\#.Plane &Loc.Embed &FID $\downarrow$ &KID $\downarrow$ &Depth $\downarrow$ \\
\hline
&4 &Linear &12.85 &8.81 &0.098 \\
&4 &Frequency &11.52 &$\boldsymbol{7.04}$ &0.110 \\
&8 &Frequency &$\boldsymbol{10.87}$ &7.29 &$\boldsymbol{0.082}$\\
\hline
\end{tabular}}
\end{center}
\caption{\textbf{Ablating the number of planes and location embeddings.} Measured in terms of FID, KID and Depth scores on SHHQ.}
\label{ab1}
\end{table}

\begin{table}
\tiny
\begin{center}
\resizebox{\linewidth}{!}{
\begin{tabular}{c c c c}
\hline
  &FID$\downarrow$ &T.C &Extra T.C \\
\hline
EG3D &4.70 &68 &-\\
Ours  (From Scratch) &4.11 &68 &-\\
EG3D + 8 plane &4.39 &- &10\\
EG3D + 12 plane &4.01 &- &10\\
\hline
\end{tabular}}
\end{center}
\caption{\textbf{Ablating the number of planes and location embeddings.} Measured in terms of FID and training cost (T.C) of V100 GPU days on FFHQ. EG3D + K plane denotes that we train from the checkpoint provided by EG3D.}
\label{ab2}
\end{table}

\subsection{Applications}

\noindent {\bf Single image 3D reconstruction.} Through pivotal tuning inversion (PTI) \cite{roich2022pivotal}, our model can be used to reconstruct 3D shapes from single views of objects. In this manuscript, we provide results of human body reconstruction with known pose conditions. We use off-the-shelf pose estimator PARE \cite{kocabas2021pare} to extract body-pose parameters and camera parameters. Following PTI \cite{roich2022pivotal}, we optimize the latent codes for 500 iterations, followed by fine-tuning the generator weights for an additional 500 iterations.

\begin{figure}[t]
\centering
\setlength{\abovecaptionskip}{0pt} 
\setlength{\belowcaptionskip}{0pt}
   \includegraphics[width=1.0\linewidth]{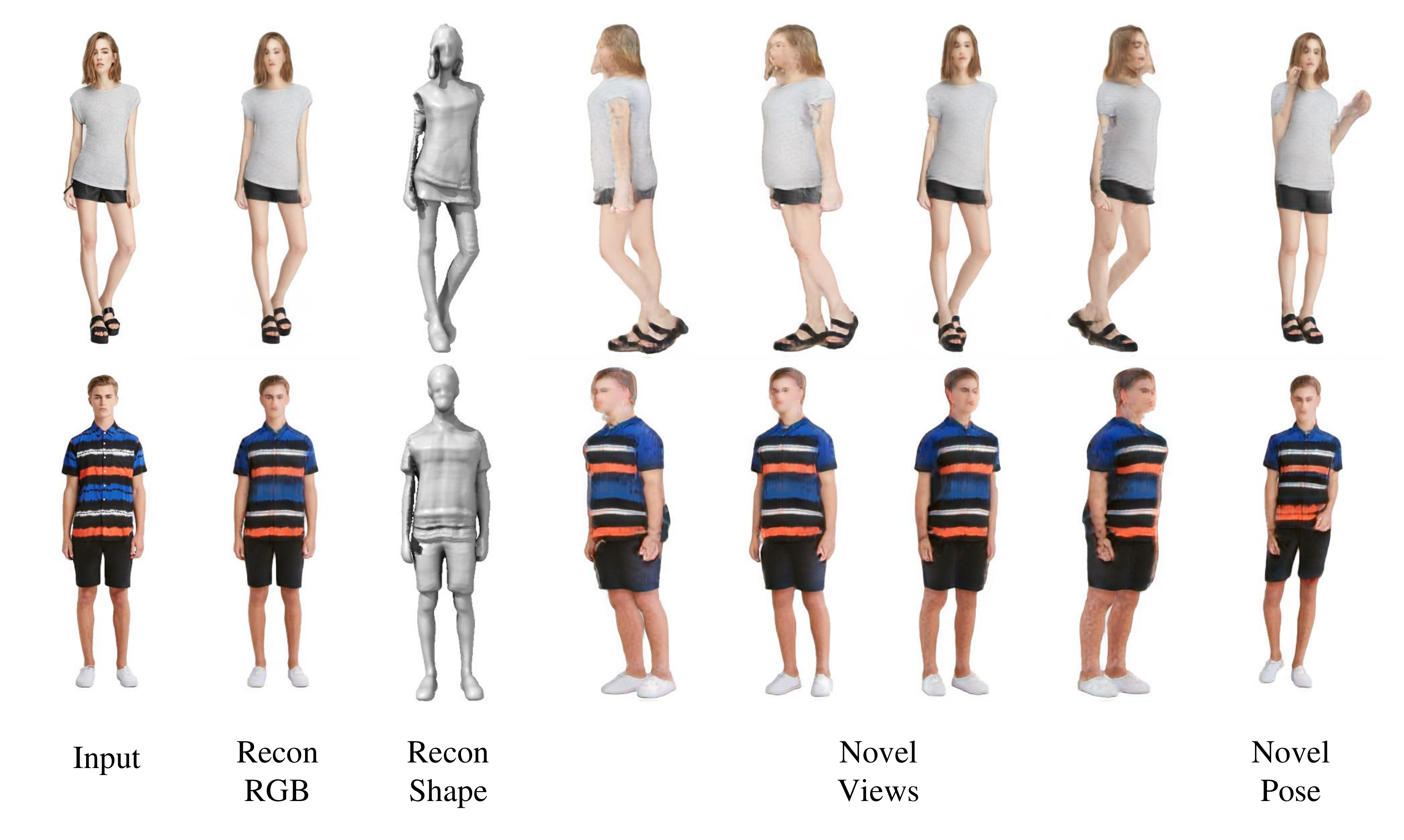}
   \caption{\textbf{Inversion on SHHQ \cite{fu2022stylegan} (Conditional).} Based on PTI \cite{roich2022pivotal}, our model can be used to reconstruct 3D shapes with single image. Furthermore, based on our conditional model, using optimized style vector and model parameters, we can synthesize images with novel view and novel pose  of the given identity.}
\label{fig:inversion_cond}
\end{figure}

As shown in Fig. \ref{fig:inversion_cond}, the reconstructed results are with fine geometry. Novel views of the same identity can also be synthesized. Furthermore, thanks to our decoupled design, we can use other poses to generate images of the same identity with novel poses, which enables the reconstructed person to be used as an animatable avatar.

\noindent {\bf Style mixing.} Our model inherits the property of StyleGAN \cite{karras2019style, karras2020analyzing, karras2021alias} latent space, although our models are all trained without style-mixing regularization. We use style-mixing in both of unconditional and pose-conditional tasks, while using the same pose as condition, the mixed samples are all with the correct pose, showing that our conditional model decouples the appearance and pose. More results and analysis are given in the supplementary.

\noindent {\bf Interpolation on latent space.} Our model generates a smooth transition between latent codes, showing that latent space learned by the model is semantically meaningful. Results are given in the supplementary materials.

\section{Discussion}

\noindent{\bf Limitations and future work.}  Although our work improves the performance of view-consistency and geometry effectively compared to previous works. There are still some artifacts at background. To tackle this problem, some modeling assumptions  \cite{zhang2020nerf++, shin2023ballgan} can be used. 

Furthermore, based on the two-stage strategy, \ie, rendering low-resolution feature map and upsampling, there are some inconsistencies of the generated results under view variation, occasionally. One possible solution is to rendering RGB image directly. To tackle the problem of the unbearable computing overhead of rendering high-resolution images, some works \cite{skorokhodov2022epigraf} introduce other strategies, where our new representation can be incorporated into.

\noindent{\bf Conclusion.} We design a novel representation orthoplanes to make 2D GAN more 3D-aware. Our model generates multi-view-consistent results and fine geometry even under extreme view angles. By introducing more explicit spatial information, our method can synthesize diverse objects and  is not confined to rigid objects. This could foster innovative methods for 3D reconstruction and facilitate the creation of 3D models for various applications.

{\small
\bibliographystyle{ieee_fullname}
\bibliography{egbib}

\begin{thebibliography}{10}\itemsep=-1pt

\bibitem{barron2022mip}
Jonathan~T Barron, Ben Mildenhall, Dor Verbin, Pratul~P Srinivasan, and Peter Hedman.
\newblock Mip-nerf 360: Unbounded anti-aliased neural radiance fields.
\newblock In {\em Proceedings of the IEEE/CVF Conference on Computer Vision and Pattern Recognition}, pages 5470--5479, 2022.

\bibitem{bergman2022generative}
Alexander~W Bergman, Petr Kellnhofer, Yifan Wang, Eric~R Chan, David~B Lindell, and Gordon Wetzstein.
\newblock Generative neural articulated radiance fields.
\newblock {\em arXiv preprint arXiv:2206.14314}, 2022.

\bibitem{binkowski2018demystifying}
Miko{\l}aj Bi{\'n}kowski, Danica~J Sutherland, Michael Arbel, and Arthur Gretton.
\newblock Demystifying mmd gans.
\newblock {\em arXiv preprint arXiv:1801.01401}, 2018.

\bibitem{chabra2020deep}
Rohan Chabra, Jan~E Lenssen, Eddy Ilg, Tanner Schmidt, Julian Straub, Steven Lovegrove, and Richard Newcombe.
\newblock Deep local shapes: Learning local sdf priors for detailed 3d reconstruction.
\newblock In {\em Computer Vision--ECCV 2020: 16th European Conference, Glasgow, UK, August 23--28, 2020, Proceedings, Part XXIX 16}, pages 608--625. Springer, 2020.

\bibitem{chan2022efficient}
Eric~R Chan, Connor~Z Lin, Matthew~A Chan, Koki Nagano, Boxiao Pan, Shalini De~Mello, Orazio Gallo, Leonidas~J Guibas, Jonathan Tremblay, Sameh Khamis, et~al.
\newblock Efficient geometry-aware 3d generative adversarial networks.
\newblock In {\em Proceedings of the IEEE/CVF Conference on Computer Vision and Pattern Recognition}, pages 16123--16133, 2022.

\bibitem{chan2021pi}
Eric~R Chan, Marco Monteiro, Petr Kellnhofer, Jiajun Wu, and Gordon Wetzstein.
\newblock pi-gan: Periodic implicit generative adversarial networks for 3d-aware image synthesis.
\newblock In {\em Proceedings of the IEEE/CVF conference on computer vision and pattern recognition}, pages 5799--5809, 2021.

\bibitem{chen2021learning}
Yinbo Chen, Sifei Liu, and Xiaolong Wang.
\newblock Learning continuous image representation with local implicit image function.
\newblock In {\em Proceedings of the IEEE/CVF conference on computer vision and pattern recognition}, pages 8628--8638, 2021.

\bibitem{choi2020stargan}
Yunjey Choi, Youngjung Uh, Jaejun Yoo, and Jung-Woo Ha.
\newblock Stargan v2: Diverse image synthesis for multiple domains.
\newblock In {\em Proceedings of the IEEE/CVF conference on computer vision and pattern recognition}, pages 8188--8197, 2020.

\bibitem{collet2015high}
Alvaro Collet, Ming Chuang, Pat Sweeney, Don Gillett, Dennis Evseev, David Calabrese, Hugues Hoppe, Adam Kirk, and Steve Sullivan.
\newblock High-quality streamable free-viewpoint video.
\newblock {\em ACM Transactions on Graphics (ToG)}, 34(4):1--13, 2015.

\bibitem{deng2019arcface}
Jiankang Deng, Jia Guo, Niannan Xue, and Stefanos Zafeiriou.
\newblock Arcface: Additive angular margin loss for deep face recognition.
\newblock In {\em Proceedings of the IEEE/CVF conference on computer vision and pattern recognition}, pages 4690--4699, 2019.

\bibitem{deng2019accurate}
Yu Deng, Jiaolong Yang, Sicheng Xu, Dong Chen, Yunde Jia, and Xin Tong.
\newblock Accurate 3d face reconstruction with weakly-supervised learning: From single image to image set.
\newblock In {\em Proceedings of the IEEE/CVF conference on computer vision and pattern recognition workshops}, pages 0--0, 2019.

\bibitem{dou2016fusion4d}
Mingsong Dou, Sameh Khamis, Yury Degtyarev, Philip Davidson, Sean~Ryan Fanello, Adarsh Kowdle, Sergio~Orts Escolano, Christoph Rhemann, David Kim, Jonathan Taylor, et~al.
\newblock Fusion4d: Real-time performance capture of challenging scenes.
\newblock {\em ACM Transactions on Graphics (ToG)}, 35(4):1--13, 2016.

\bibitem{fridovich2023k}
Sara Fridovich-Keil, Giacomo Meanti, Frederik Warburg, Benjamin Recht, and Angjoo Kanazawa.
\newblock K-planes: Explicit radiance fields in space, time, and appearance.
\newblock {\em arXiv preprint arXiv:2301.10241}, 2023.

\bibitem{fridovich2022plenoxels}
Sara Fridovich-Keil, Alex Yu, Matthew Tancik, Qinhong Chen, Benjamin Recht, and Angjoo Kanazawa.
\newblock Plenoxels: Radiance fields without neural networks.
\newblock In {\em Proceedings of the IEEE/CVF Conference on Computer Vision and Pattern Recognition}, pages 5501--5510, 2022.

\bibitem{fu2022stylegan}
Jianglin Fu, Shikai Li, Yuming Jiang, Kwan-Yee Lin, Chen Qian, Chen~Change Loy, Wayne Wu, and Ziwei Liu.
\newblock Stylegan-human: A data-centric odyssey of human generation.
\newblock In {\em Computer Vision--ECCV 2022: 17th European Conference, Tel Aviv, Israel, October 23--27, 2022, Proceedings, Part XVI}, pages 1--19. Springer, 2022.

\bibitem{gadelha20173d}
Matheus Gadelha, Subhransu Maji, and Rui Wang.
\newblock 3d shape induction from 2d views of multiple objects.
\newblock In {\em 2017 International Conference on 3D Vision (3DV)}, pages 402--411. IEEE, 2017.

\bibitem{goddard2018ucsf}
Thomas~D Goddard, Conrad~C Huang, Elaine~C Meng, Eric~F Pettersen, Gregory~S Couch, John~H Morris, and Thomas~E Ferrin.
\newblock Ucsf chimerax: Meeting modern challenges in visualization and analysis.
\newblock {\em Protein Science}, 27(1):14--25, 2018.

\bibitem{goodfellow2020generative}
Ian Goodfellow, Jean Pouget-Abadie, Mehdi Mirza, Bing Xu, David Warde-Farley, Sherjil Ozair, Aaron Courville, and Yoshua Bengio.
\newblock Generative adversarial networks.
\newblock {\em Communications of the ACM}, 63(11):139--144, 2020.

\bibitem{gropp2020implicit}
Amos Gropp, Lior Yariv, Niv Haim, Matan Atzmon, and Yaron Lipman.
\newblock Implicit geometric regularization for learning shapes.
\newblock {\em arXiv preprint arXiv:2002.10099}, 2020.

\bibitem{gu2021stylenerf}
Jiatao Gu, Lingjie Liu, Peng Wang, and Christian Theobalt.
\newblock Stylenerf: A style-based 3d-aware generator for high-resolution image synthesis.
\newblock {\em arXiv preprint arXiv:2110.08985}, 2021.

\bibitem{hedman2021baking}
Peter Hedman, Pratul~P Srinivasan, Ben Mildenhall, Jonathan~T Barron, and Paul Debevec.
\newblock Baking neural radiance fields for real-time view synthesis.
\newblock In {\em Proceedings of the IEEE/CVF International Conference on Computer Vision}, pages 5875--5884, 2021.

\bibitem{henzler2019escaping}
Philipp Henzler, Niloy~J Mitra, and Tobias Ritschel.
\newblock Escaping plato's cave: 3d shape from adversarial rendering.
\newblock In {\em Proceedings of the IEEE/CVF International Conference on Computer Vision}, pages 9984--9993, 2019.

\bibitem{heusel2017gans}
Martin Heusel, Hubert Ramsauer, Thomas Unterthiner, Bernhard Nessler, and Sepp Hochreiter.
\newblock Gans trained by a two time-scale update rule converge to a local nash equilibrium.
\newblock {\em Advances in neural information processing systems}, 30, 2017.

\bibitem{hong2022eva3d}
Fangzhou Hong, Zhaoxi Chen, Yushi Lan, Liang Pan, and Ziwei Liu.
\newblock Eva3d: Compositional 3d human generation from 2d image collections.
\newblock {\em arXiv preprint arXiv:2210.04888}, 2022.

\bibitem{jacobson2012fast}
Alec Jacobson, Ilya Baran, Ladislav Kavan, Jovan Popovi{\'c}, and Olga Sorkine.
\newblock Fast automatic skinning transformations.
\newblock {\em ACM Transactions on Graphics (TOG)}, 31(4):1--10, 2012.

\bibitem{jiang2020local}
Chiyu Jiang, Avneesh Sud, Ameesh Makadia, Jingwei Huang, Matthias Nie{\ss}ner, Thomas Funkhouser, et~al.
\newblock Local implicit grid representations for 3d scenes.
\newblock In {\em Proceedings of the IEEE/CVF Conference on Computer Vision and Pattern Recognition}, pages 6001--6010, 2020.

\bibitem{jiang2020sdfdiff}
Yue Jiang, Dantong Ji, Zhizhong Han, and Matthias Zwicker.
\newblock Sdfdiff: Differentiable rendering of signed distance fields for 3d shape optimization.
\newblock In {\em Proceedings of the IEEE/CVF Conference on Computer Vision and Pattern Recognition}, pages 1251--1261, 2020.

\bibitem{karras2020training}
Tero Karras, Miika Aittala, Janne Hellsten, Samuli Laine, Jaakko Lehtinen, and Timo Aila.
\newblock Training generative adversarial networks with limited data.
\newblock {\em Advances in neural information processing systems}, 33:12104--12114, 2020.

\bibitem{karras2021alias}
Tero Karras, Miika Aittala, Samuli Laine, Erik H{\"a}rk{\"o}nen, Janne Hellsten, Jaakko Lehtinen, and Timo Aila.
\newblock Alias-free generative adversarial networks.
\newblock {\em Advances in Neural Information Processing Systems}, 34:852--863, 2021.

\bibitem{karras2019style}
Tero Karras, Samuli Laine, and Timo Aila.
\newblock A style-based generator architecture for generative adversarial networks.
\newblock In {\em Proceedings of the IEEE/CVF conference on computer vision and pattern recognition}, pages 4401--4410, 2019.

\bibitem{karras2020analyzing}
Tero Karras, Samuli Laine, Miika Aittala, Janne Hellsten, Jaakko Lehtinen, and Timo Aila.
\newblock Analyzing and improving the image quality of stylegan.
\newblock In {\em Proceedings of the IEEE/CVF conference on computer vision and pattern recognition}, pages 8110--8119, 2020.

\bibitem{kellnhofer2021neural}
Petr Kellnhofer, Lars~C Jebe, Andrew Jones, Ryan Spicer, Kari Pulli, and Gordon Wetzstein.
\newblock Neural lumigraph rendering.
\newblock In {\em Proceedings of the IEEE/CVF Conference on Computer Vision and Pattern Recognition}, pages 4287--4297, 2021.

\bibitem{kocabas2021pare}
Muhammed Kocabas, Chun-Hao~P Huang, Otmar Hilliges, and Michael~J Black.
\newblock Pare: Part attention regressor for 3d human body estimation.
\newblock In {\em Proceedings of the IEEE/CVF International Conference on Computer Vision}, pages 11127--11137, 2021.

\bibitem{lindell2021autoint}
David~B Lindell, Julien~NP Martel, and Gordon Wetzstein.
\newblock Autoint: Automatic integration for fast neural volume rendering.
\newblock In {\em Proceedings of the IEEE/CVF Conference on Computer Vision and Pattern Recognition}, pages 14556--14565, 2021.

\bibitem{liu2020neural}
Lingjie Liu, Jiatao Gu, Kyaw Zaw~Lin, Tat-Seng Chua, and Christian Theobalt.
\newblock Neural sparse voxel fields.
\newblock {\em Advances in Neural Information Processing Systems}, 33:15651--15663, 2020.

\bibitem{loper2015smpl}
Matthew Loper, Naureen Mahmood, Javier Romero, Gerard Pons-Moll, and Michael~J Black.
\newblock Smpl: A skinned multi-person linear model.
\newblock {\em ACM transactions on graphics (TOG)}, 34(6):1--16, 2015.

\bibitem{martel2021acorn}
Julien~NP Martel, David~B Lindell, Connor~Z Lin, Eric~R Chan, Marco Monteiro, and Gordon Wetzstein.
\newblock Acorn: Adaptive coordinate networks for neural scene representation.
\newblock {\em arXiv preprint arXiv:2105.02788}, 2021.

\bibitem{max1995optical}
Nelson Max.
\newblock Optical models for direct volume rendering.
\newblock {\em IEEE Transactions on Visualization and Computer Graphics}, 1(2):99--108, 1995.

\bibitem{mescheder2018training}
Lars Mescheder, Andreas Geiger, and Sebastian Nowozin.
\newblock Which training methods for gans do actually converge?
\newblock In {\em International conference on machine learning}, pages 3481--3490. PMLR, 2018.

\bibitem{mildenhall2020nerf}
B Mildenhall, PP Srinivasan, M Tancik, JT Barron, R Ramamoorthi, and R Ng.
\newblock Nerf: Representing scenes as neural radiance fields for view synthesis.
\newblock In {\em European conference on computer vision}, 2020.

\bibitem{muller2022instant}
Thomas M{\"u}ller, Alex Evans, Christoph Schied, and Alexander Keller.
\newblock Instant neural graphics primitives with a multiresolution hash encoding.
\newblock {\em ACM Transactions on Graphics (ToG)}, 41(4):1--15, 2022.

\bibitem{nguyen2019hologan}
Thu Nguyen-Phuoc, Chuan Li, Lucas Theis, Christian Richardt, and Yong-Liang Yang.
\newblock Hologan: Unsupervised learning of 3d representations from natural images.
\newblock In {\em Proceedings of the IEEE/CVF International Conference on Computer Vision}, pages 7588--7597, 2019.

\bibitem{nguyen2020blockgan}
Thu~H Nguyen-Phuoc, Christian Richardt, Long Mai, Yongliang Yang, and Niloy Mitra.
\newblock Blockgan: Learning 3d object-aware scene representations from unlabelled images.
\newblock {\em Advances in neural information processing systems}, 33:6767--6778, 2020.

\bibitem{niemeyer2021giraffe}
Michael Niemeyer and Andreas Geiger.
\newblock Giraffe: Representing scenes as compositional generative neural feature fields.
\newblock In {\em Proceedings of the IEEE/CVF Conference on Computer Vision and Pattern Recognition}, pages 11453--11464, 2021.

\bibitem{noguchi2022unsupervised}
Atsuhiro Noguchi, Xiao Sun, Stephen Lin, and Tatsuya Harada.
\newblock Unsupervised learning of efficient geometry-aware neural articulated representations.
\newblock In {\em Computer Vision--ECCV 2022: 17th European Conference, Tel Aviv, Israel, October 23--27, 2022, Proceedings, Part XVII}, pages 597--614. Springer, 2022.

\bibitem{or2022stylesdf}
Roy Or-El, Xuan Luo, Mengyi Shan, Eli Shechtman, Jeong~Joon Park, and Ira Kemelmacher-Shlizerman.
\newblock Stylesdf: High-resolution 3d-consistent image and geometry generation.
\newblock In {\em Proceedings of the IEEE/CVF Conference on Computer Vision and Pattern Recognition}, pages 13503--13513, 2022.

\bibitem{park2019deepsdf}
Jeong~Joon Park, Peter Florence, Julian Straub, Richard Newcombe, and Steven Lovegrove.
\newblock Deepsdf: Learning continuous signed distance functions for shape representation.
\newblock In {\em Proceedings of the IEEE/CVF conference on computer vision and pattern recognition}, pages 165--174, 2019.

\bibitem{peng2021neural}
Sida Peng, Yuanqing Zhang, Yinghao Xu, Qianqian Wang, Qing Shuai, Hujun Bao, and Xiaowei Zhou.
\newblock Neural body: Implicit neural representations with structured latent codes for novel view synthesis of dynamic humans.
\newblock In {\em Proceedings of the IEEE/CVF Conference on Computer Vision and Pattern Recognition}, pages 9054--9063, 2021.

\bibitem{ranftl2020towards}
Ren{\'e} Ranftl, Katrin Lasinger, David Hafner, Konrad Schindler, and Vladlen Koltun.
\newblock Towards robust monocular depth estimation: Mixing datasets for zero-shot cross-dataset transfer.
\newblock {\em IEEE transactions on pattern analysis and machine intelligence}, 44(3):1623--1637, 2020.

\bibitem{reiser2021kilonerf}
Christian Reiser, Songyou Peng, Yiyi Liao, and Andreas Geiger.
\newblock Kilonerf: Speeding up neural radiance fields with thousands of tiny mlps.
\newblock In {\em Proceedings of the IEEE/CVF International Conference on Computer Vision}, pages 14335--14345, 2021.

\bibitem{roich2022pivotal}
Daniel Roich, Ron Mokady, Amit~H Bermano, and Daniel Cohen-Or.
\newblock Pivotal tuning for latent-based editing of real images.
\newblock {\em ACM Transactions on Graphics (TOG)}, 42(1):1--13, 2022.

\bibitem{schwarz2020graf}
Katja Schwarz, Yiyi Liao, Michael Niemeyer, and Andreas Geiger.
\newblock Graf: Generative radiance fields for 3d-aware image synthesis.
\newblock {\em Advances in Neural Information Processing Systems}, 33:20154--20166, 2020.

\bibitem{schwarz2022voxgraf}
Katja Schwarz, Axel Sauer, Michael Niemeyer, Yiyi Liao, and Andreas Geiger.
\newblock Voxgraf: Fast 3d-aware image synthesis with sparse voxel grids.
\newblock {\em arXiv preprint arXiv:2206.07695}, 2022.

\bibitem{shi2023learning}
Zifan Shi, Yujun Shen, Yinghao Xu, Sida Peng, Yiyi Liao, Sheng Guo, Qifeng Chen, and Dit-Yan Yeung.
\newblock Learning 3d-aware image synthesis with unknown pose distribution.
\newblock {\em arXiv preprint arXiv:2301.07702}, 2023.

\bibitem{shin2023ballgan}
Minjung Shin, Yunji Seo, Jeongmin Bae, Young~Sun Choi, Hyunsu Kim, Hyeran Byun, and Youngjung Uh.
\newblock Ballgan: 3d-aware image synthesis with a spherical background.
\newblock {\em arXiv preprint arXiv:2301.09091}, 2023.

\bibitem{sitzmann2019deepvoxels}
Vincent Sitzmann, Justus Thies, Felix Heide, Matthias Nie{\ss}ner, Gordon Wetzstein, and Michael Zollhofer.
\newblock Deepvoxels: Learning persistent 3d feature embeddings.
\newblock In {\em Proceedings of the IEEE/CVF Conference on Computer Vision and Pattern Recognition}, pages 2437--2446, 2019.

\bibitem{skorokhodov2022epigraf}
Ivan Skorokhodov, Sergey Tulyakov, Yiqun Wang, and Peter Wonka.
\newblock Epigraf: Rethinking training of 3d gans.
\newblock {\em arXiv preprint arXiv:2206.10535}, 2022.

\bibitem{tancik2022block}
Matthew Tancik, Vincent Casser, Xinchen Yan, Sabeek Pradhan, Ben Mildenhall, Pratul~P Srinivasan, Jonathan~T Barron, and Henrik Kretzschmar.
\newblock Block-nerf: Scalable large scene neural view synthesis.
\newblock In {\em Proceedings of the IEEE/CVF Conference on Computer Vision and Pattern Recognition}, pages 8248--8258, 2022.

\bibitem{tewari2020state}
Ayush Tewari, Ohad Fried, Justus Thies, Vincent Sitzmann, Stephen Lombardi, Kalyan Sunkavalli, Ricardo Martin-Brualla, Tomas Simon, Jason Saragih, Matthias Nie{\ss}ner, et~al.
\newblock State of the art on neural rendering.
\newblock In {\em Computer Graphics Forum}, volume~39, pages 701--727. Wiley Online Library, 2020.

\bibitem{weng2022humannerf}
Chung-Yi Weng, Brian Curless, Pratul~P Srinivasan, Jonathan~T Barron, and Ira Kemelmacher-Shlizerman.
\newblock Humannerf: Free-viewpoint rendering of moving people from monocular video.
\newblock In {\em Proceedings of the IEEE/CVF Conference on Computer Vision and Pattern Recognition}, pages 16210--16220, 2022.

\bibitem{xu2022pv3d}
Eric~Zhongcong Xu, Jianfeng Zhang, Jun~Hao Liew, Wenqing Zhang, Song Bai, Jiashi Feng, and Mike~Zheng Shou.
\newblock Pv3d: A 3d generative model for portrait video generation.
\newblock {\em arXiv preprint arXiv:2212.06384}, 2022.

\bibitem{yang20223dhumangan}
Zhuoqian Yang, Shikai Li, Wayne Wu, and Bo Dai.
\newblock 3dhumangan: Towards photo-realistic 3d-aware human image generation.
\newblock {\em arXiv preprint arXiv:2212.07378}, 2022.

\bibitem{yariv2021volume}
Lior Yariv, Jiatao Gu, Yoni Kasten, and Yaron Lipman.
\newblock Volume rendering of neural implicit surfaces.
\newblock {\em Advances in Neural Information Processing Systems}, 34:4805--4815, 2021.

\bibitem{yifan2021geometry}
Wang Yifan, Lukas Rahmann, and Olga Sorkine-Hornung.
\newblock Geometry-consistent neural shape representation with implicit displacement fields.
\newblock {\em arXiv preprint arXiv:2106.05187}, 2021.

\bibitem{zhang2022avatargen}
Jianfeng Zhang, Zihang Jiang, Dingdong Yang, Hongyi Xu, Yichun Shi, Guoxian Song, Zhongcong Xu, Xinchao Wang, and Jiashi Feng.
\newblock Avatargen: a 3d generative model for animatable human avatars.
\newblock {\em arXiv preprint arXiv:2211.14589}, 2022.

\bibitem{zhang2020nerf++}
Kai Zhang, Gernot Riegler, Noah Snavely, and Vladlen Koltun.
\newblock Nerf++: Analyzing and improving neural radiance fields.
\newblock {\em arXiv preprint arXiv:2010.07492}, 2020.

\bibitem{zhao2022generative}
Xiaoming Zhao, Fangchang Ma, David G{\"u}era, Zhile Ren, Alexander~G Schwing, and Alex Colburn.
\newblock Generative multiplane images: Making a 2d gan 3d-aware.
\newblock In {\em Computer Vision--ECCV 2022: 17th European Conference, Tel Aviv, Israel, October 23--27, 2022, Proceedings, Part V}, pages 18--35. Springer, 2022.

\bibitem{zhou2021cips}
Peng Zhou, Lingxi Xie, Bingbing Ni, and Qi Tian.
\newblock Cips-3d: A 3d-aware generator of gans based on conditionally-independent pixel synthesis.
\newblock {\em arXiv preprint arXiv:2110.09788}, 2021.

\end{thebibliography}
}

\clearpage
\appendix



\textbf{\LARGE Appendix}
\vspace{5ex}

In the appendix, we first provide implementation details including architectures of our generators and hyperparameters of all experiments (Sec. \ref{impd}). Additionally, there are more experiments to demonstrate the effectiveness of proposed method (Sec. \ref{ade}). Finally, more visual results including self-evaluation, comparison and applications are provided (Sec. \ref{avr}). For more visual results, please view the accompanying supplemental video.

\section{Implementation details}
\label{impd}

\subsection{Unconditional GAN}

For unconditional GAN, we implemented our framework on top of the official PyTorch implementation of EG3D\footnote{\url{https://github.com/NVlabs/eg3d}} \cite{chan2022efficient}. 

\noindent{\bf Backbone.} Our backbone, \ie, OrthoPlanes generator is with a mapping network of 2 hidden layers for all experiments on unconditional tasks. The backbone operates at a resolution of $256^2$ with channel of 96. For the majority of our experiments, we use orthoplanes of $K=8$ (\ie, $3K=24$ planes in total.) with frequency encoding. For large datasets, such as FFHQ \cite{karras2019style} and SHHQ \cite{fu2022stylegan}, we train with random initialization; for small datasets AFHQv2-Cat \cite{choi2020stargan}, we follow previous works \cite{karras2020training, chan2022efficient} by transfer learning from a checkpoint trained on a larger dataset. For our method on AFHQv2-Cat, we use the checkpoint trained on FFHQ, provided by EG3D \cite{chan2022efficient} with our new branch randomly initialized.

The implementation of our ToFeature Branch, which used to generate orhtoplanes efficiently, is based on the ToRGB layers in StyleGAN2 \cite{karras2020analyzing} with a affine layer which maps the concatenation of style vector and location embedding to the vector used to modulate the weights of convolutional kernels in the synthesis block.

Based on such implementaion, our generator is scalable on the number of planes during training. To demonstrate the scalability of our representation, we train our model from checkpoint provided by EG3D \cite{chan2022efficient}, which can be regarded as orthoplanes with $K=1$. With extra 0.14 times EG3D full-pipeline-training time,  our approach demonstrates significant improvements in FID scores.

\noindent{\bf R1 Regularization.} We use R1 regularizaton \cite{mescheder2018training} with $\gamma=1$ for FFHQ \cite{karras2019style}, $\gamma=7.8$ for AFHQv2-Cat \cite{choi2020stargan} and $\gamma=5$ for SHHQ \cite{fu2022stylegan}.

\noindent{\bf Other configurations.} Other configurations not mentioned are as the same as the configurations in EG3D \cite{chan2022efficient}.

\noindent{\bf Training.} We train all models with a batch size of 32. The learning rates are 2.5e-3 and 2e-3 for generator and discriminator, respectively. Following \cite{karras2021alias, chan2022efficient}, we blur the first 200K images entering the discriminator.

For experiments on FFHQ \cite{karras2019style} and AFHQv2-Cat \cite{choi2020stargan}, we use the two-stage training strategy introduced in EG3D \cite{chan2022efficient}, we train at a resolution of $64^2$ for 12M images and at $128^2$ for an additional 2M images for FFHQ \cite{karras2019style} and 6M + 2M for AFHQv2-Cat \cite{choi2020stargan}. For experiments on SHHQ \cite{fu2022stylegan}, we train at a resolution of $64^2$ for 5M images, without additional high-resolution training.

For experiments on AFHQv2-Cat \cite{choi2020stargan}, we use the adaptive discriminator augmentation \cite{karras2020analyzing} additionally.
\begin{figure*}
    \centering
    \vspace{-2ex}
    \includegraphics[width=0.95\textwidth,height=45mm]{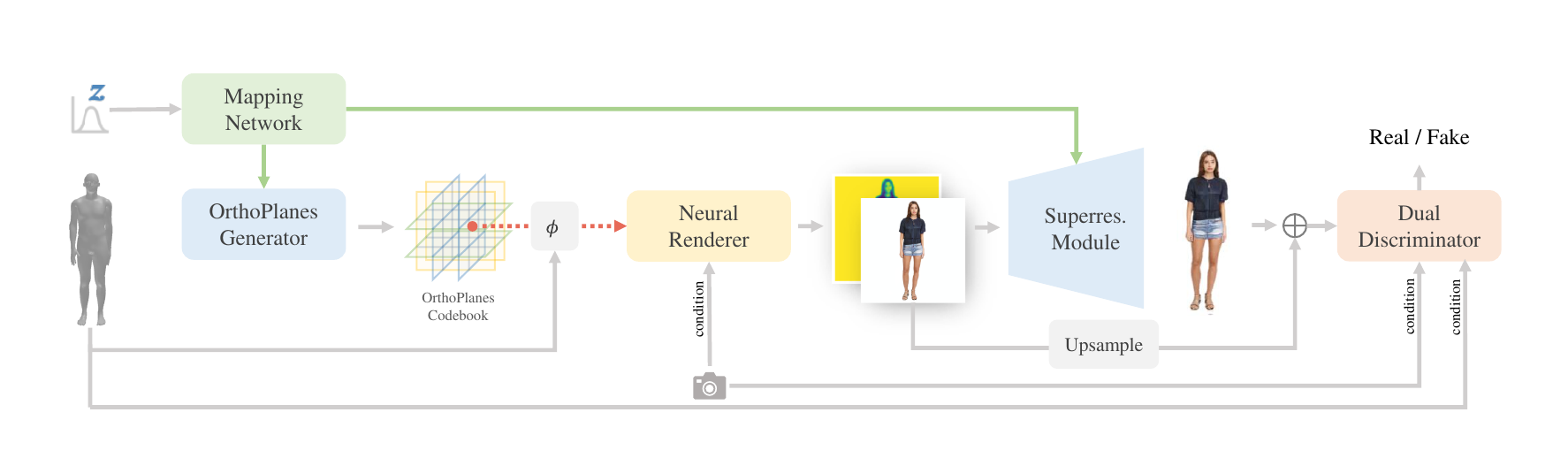}
    \vspace{-1ex}
    \captionof{figure}{\textbf{Our conditional GAN framework.} Our generator is comprised by several parts: a pose-canonical-space codebook generator with a mapping network, a point-wise warping network $\phi$, a neural renderer implemented by {\bf orhtoplanes} representation with a lightweight MLP and a volumetric renderer and a super resolution module. The discriminator is a dual-discriminator to avoid view-inconsistency or pose-inconsistency.}
    \vspace{-1ex}
    
\label{fig:cond_pipe}
\end{figure*}

\subsection{Conditional GAN}

To further demonstrate the effectiveness of our representation, especially for articulated objects with the problem of pose-consistency, we do additional experiments on pose-controllable avatar synthesis. For conditional GAN, Fig. \ref{fig:cond_pipe} gives an overview of our architecture. In the task of pose-conditional human body image synthesis, the pose of generated image should be consistent with the given condition. 

In our architecture, the pose parameters $\theta$ defined by SMPL \cite{loper2015smpl} and corresponding mesh will be used as the condition for both of generator and discriminator. The output of generator is pose-dependent clothed human image with reasonable geometry. To tackle the problems of view-consistency and pose-consistency mixed together, we use the strategy of generating canonical human-body used in other works \cite{peng2021neural, weng2022humannerf, bergman2022generative, hong2022eva3d, zhang2022avatargen}.

\noindent{\bf Pose-Canonical Human Generation.} We leverage the

\noindent orthoplanes representation for generation of clothed humans in the canonical space $\mathcal{C}$. Specifically, the random latent code $\boldsymbol{z}$ will be used to generate the canonical feature of clothed humans via CodeBook Generator. Every point sampled in observation space $\mathcal{O}$ will be transformed to this space, and query the orthoplanes coodbook to get its feature. Note that the generator only  needs to generate information of the human in canonical space, which means that the generator only needs to tackle the problem of view-consistency, alleviating the difficulty of training it from in-the-wild 2D images.

\noindent{\bf Pose-Guided Transformation.} To transform the point $\boldsymbol{x}$ sampled in the observed space $\mathcal{O}$ to the pose-canonical sapce $\mathcal{C}$ under given SMPL \cite{loper2015smpl} pose $\boldsymbol{\theta}$, we use inverse LBS \cite{jacobson2012fast} to get its location $\boldsymbol{x}_c^*$ in pose-canonical space. Specifically, we denote the human body vertices defined by the SMPL under given pose $\boldsymbol{\theta}$ as $\mathcal{V}=\left\{v_i\right\}_{i=1:N}$ ($N=6890$ in SMPL format), we can obtain basic warped point $\boldsymbol{x}_c^*$ as $\boldsymbol{x}_c^*=(\sum_{k=1}^{K} w_k(v_p^*)G_k(\theta))^{-1}\boldsymbol{x}$, where $v_p^*\in \mathcal{V}$ is the nearest vertex to the querying point $\boldsymbol{x}$; $\left\{w_k(v_p^*)\right\}$ satisfies $\sum_{k=1}^{K}w_k(v_p^*)=1, w_k(v_p^*)\geq0$ are blend weights of the vertex; $G_k(\theta)\in SE(3)$ denotes the cumulative linear transformation at the $k$-th skeleton joint defined by SMPL. 

In our practice, we found that naively using inverse LBS would cause the training unstable, since the representation is a codebook indeed, and points far from the mesh will be warped to inaccurate locations, causing collision among points. To tackle this problem, we use a simple but useful method, \ie, we only do the inverse LBS on points within the distance $d_r$ from the mesh, and for other points, we just transform them using the root rotation defined by SMPL

\begin{equation}
    \label{transform}
         \boldsymbol{x}_c^*  = \left\{
    \begin{array}{rcl}
         (\sum_{k=1}^{K} w_k(v_p^*)G_k(\theta))^{-1}
         \boldsymbol{x}   &d(x,\mathcal{V})\leq d_r  \\
         G_1^{-1}(\theta)
         \boldsymbol{x}   &otherwise
    \end{array}
    \right .
\end{equation}

\noindent{\bf Point-Wise Deformation.} Although pose-guided inverse LBS can transform any spatial points to the pose-canonical space, it only considers the rigid motion, lack of pose-dependent deformation, \eg,  cloth wrinkles. We introduce a SIREN \cite{chan2021pi} based encoder to describe the non-rigid transformation of the canonical point $\boldsymbol{x}_c^*$ under given pose parameter $\boldsymbol{\theta}$ and appearance code $\boldsymbol{w}$, the canonical point will be added small high-frequency offset, where $\alpha_d$ is set empirically to limit the scalar of the offset $\Delta x_d$. Finally, the point $\boldsymbol{x}$ in the observed space will be transformed to $\boldsymbol{x}^*=\boldsymbol{x}_c^*+\alpha_d\cdot \boldsymbol{\Delta x_d}$, and then query the canonical ortho-planes to get its feature. Specifically, out deformation module is composited of three siren blocks, each with frequency and bias mapping layers (2-layer MLP with Softplus activation functions).

The point-wise warping network $\phi$, as shown in Fig. \ref{fig:cond_pipe}, is the composite function of pose-guided LBS and point-wise deformation.

\noindent{\bf Delta SDF Learning Strategy.} In-the-wild human-body 2D image collections, are usually with very imbalance pose and view distribution, making it hard for a network to learn correct canonical 3D human without any explicit supervisory. Like concurrent works \cite{hong2022eva3d, zhang2022avatargen}, we use the Delta SDF \cite{yifan2021geometry} learning strategy instead of predicting density directly. Specifically, the spatial point $\boldsymbol{x}$ will first compute its signed distance $d_0(\boldsymbol{x})$ from the SMPL Template mesh under given pose $\boldsymbol{\theta}$, instead of output the density $\sigma(\boldsymbol{x})$, we just output the offset $\Delta d(\boldsymbol{x})$ from $d_0(\boldsymbol{x})$ \cite{yifan2021geometry}, and then translate SDF value to the density value using Eq. \eqref{sdf2den}.

\begin{figure*}
    \centering
    \vspace{-2ex}
    \includegraphics[width=0.95\textwidth,height=118mm]{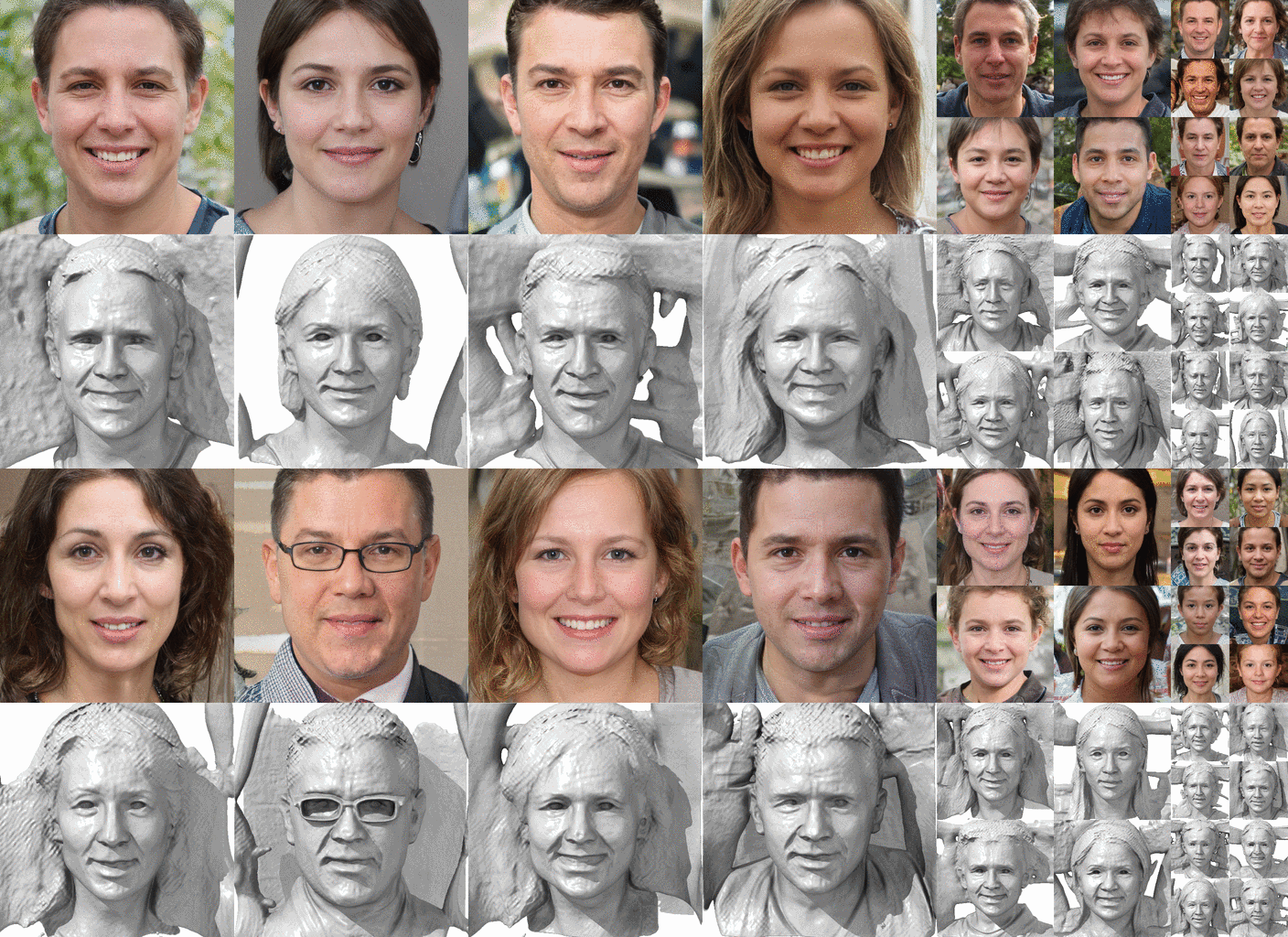}
    \vspace{-1ex}
    \captionof{figure}{\textbf{Images and geometry for seeds 0-31 on FFHQ \cite{karras2019style} $512^2$.} Sampled with truncation \cite{karras2019style} $\psi=0.5$.}
    \vspace{-1ex}
    
\label{fig:ffhq}
\end{figure*}

\begin{equation}
    (\boldsymbol{c},\Delta d)(\boldsymbol{x})=MLP(\boldsymbol{F}(\boldsymbol{x}))
\end{equation}

\begin{equation}
\label{sdf2den}
    \sigma(\boldsymbol{x})=\frac{1}{\alpha}\cdot Sigmoid(-\frac{d_0(x)+\Delta d(x)}{\alpha})
\end{equation}

where $\boldsymbol{c}$ is the feature vector of the point $\boldsymbol{x}$, $\alpha > 0$ is a learnable parameter. Both of $\boldsymbol{c}$ and  $\sigma$ will be processed by a neural volume renderer to get the 2D feature image.

\noindent{\bf Volume Renderer and Super Resolution.} Volume renderer and super resolution module are the same as in

\noindent unconditional experiments. In conditional experiments, we do not use the two-pass importance sampling strategy to sample points in observation space but only uniform sampling.
\noindent{\bf Training.} In our conditional GAN experiments, we also use non-saturating GAN loss \cite{goodfellow2020generative} with R1 regularization \cite{mescheder2018training} for training. For the discriminator, We use the same architecture as dual-discriminator proposed by EG3D \cite{chan2022efficient}, with additional pose parameters $\boldsymbol{\theta}$ as condition. Furthermore, based on methods mentioned before, there are some additional regularization terms  like Eikonal loss \cite{gropp2020implicit}. All regularization terms and weights are given in supplement.

\noindent{\bf Training.} To train our model end-to-end, we use non-saturating GAN loss \cite{goodfellow2020generative} with R1 regularization \cite{mescheder2018training}. For the discriminator, We use the same architecture as dual-discriminator proposed by EG3D \cite{chan2022efficient}, with additional pose parameters $\boldsymbol{\theta}$ as condition. So the adversarial loss is

\begin{align}
     \mathcal{L}_{adv,D} & = \mathbf{E}_{\boldsymbol{z}\sim P_z, \{\boldsymbol{\theta}, \boldsymbol{c}\}\sim P_{data}} [f(D(G(\boldsymbol{z};\boldsymbol{\theta},\boldsymbol{c}))) ] \\ & + \mathbf{E}_{\boldsymbol{I}\sim P_{data}} [f(D(\boldsymbol{I})) + \frac{\gamma}{2} ||\nabla_{\boldsymbol{I}}D(\boldsymbol{I})||^2_2] \\
     \mathcal{L}_{adv,G} &= \mathbf{E}_{\boldsymbol{z}\sim P_z, \{\boldsymbol{\theta}, \boldsymbol{c}\}\sim P_{data}} [f(-D(G(\boldsymbol{z};\boldsymbol{\theta},\boldsymbol{c}))) ]
\end{align}

where $f(x)=-\log(1+\exp{-x})$; $\boldsymbol{\theta}, \boldsymbol{c}$ are pose and view conditions estimated from real images, respectively. 

Other than adversarial loss, some more regularization terms for training generator are introduced. 

First, for any point, the offset caused by pose like cloth wrinkles  usually small, so we adopt a deformation regularization term $\mathcal{L}_{deform}$ like AvatarGen \cite{zhang2022avatargen}.

\begin{equation}
    \mathcal{L}_{deform}=\mathbf{E}_{\boldsymbol{x},\boldsymbol{\theta}\sim P_{data},\boldsymbol{z}\sim p_Z}[||\Delta \boldsymbol{x}_{\boldsymbol{d}}(\boldsymbol{x};\boldsymbol{\theta},\boldsymbol{z})||_2^2]
\end{equation}

Second, the minimum offset of the predicted SDF from the template mesh should be constrained to maintain plausible human geometry \cite{hong2022eva3d, zhang2022avatargen}. However, there are also some  elements with higher degrees of freedom like hairs and dresses. So we adopt a piecewise function $\mathcal{L}_{offset}$ to regularize the learned SDF.
\begin{figure*}
    \centering
    \vspace{-2ex}
    \includegraphics[width=0.95\textwidth,height=118mm]{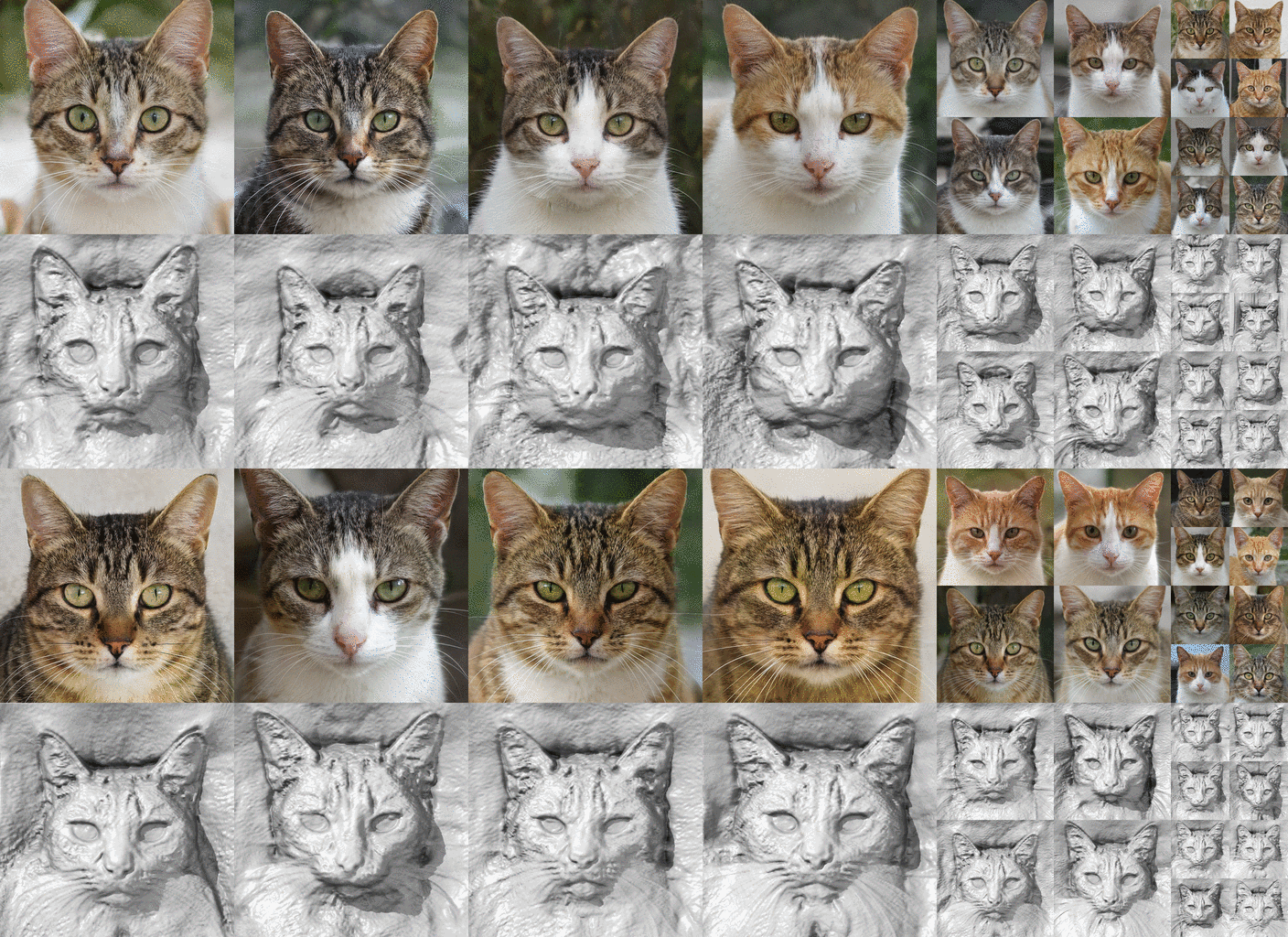}
    \vspace{-1ex}
    \captionof{figure}{\textbf{Images and geometry for seeds 0-31 on AFHQv2-Cat \cite{choi2020stargan} $512^2$.} Sampled with truncation \cite{karras2019style} $\psi=0.5$.}
    \vspace{-1ex}
    
\label{fig:afhq}
\end{figure*}

\begin{align}
     \mathcal{L}_{offset}  & =  \mathbf{E}_{\boldsymbol{x}}[w_x\cdot ||\Delta d(x)||^2_2] \\ w_x & = 
     \left\{     
     \begin{array}{rcl}         
     e^{-\frac{||d_0(x)||^2_2}{k}}   &||d_0(x)||_1\leq d_r  \\ 
     1 & otherwise
     \end{array}
     \right .
\end{align}

where $w_x$ is the weight decreasing with distance from the template mesh like \cite{zhang2022avatargen}, $k$ and $d_r$ are set empirically.

\noindent Intuitively, this term allows higher degrees of freedom for the points within a certain range.

Third, to ensure that the learned SDF is physically valid, Eikonal loss $\mathcal{L}_{eik}$ \cite{gropp2020implicit} is adopted. 

\begin{equation}
    \mathcal{L}_{eik}=\mathbf{E}_{\boldsymbol{x}}[||\nabla_{\boldsymbol{x}}d(\boldsymbol{x}) - 1||^2_2]
\end{equation}

where $\boldsymbol{x}$ and $d(\boldsymbol{x})=d_0(\boldsymbol{x}) + \Delta d(\boldsymbol{x})$ are sampled point and the predicted signed distance, respectively.

So the overall loss for training the generator $\mathcal{L}_G$ is formulated as

\begin{equation}
    \mathcal{L}_G = \mathcal{L}_{adv,G} + \lambda_1 \mathcal{L}_{deform} + \lambda_2 \mathcal{L}_{offset} + \lambda_3 \mathcal{L}_{eik}
\end{equation}

where $\lambda_{i}$ are the corresponding weights. In our experiments, we use $\lambda_1=0.5,\lambda_2=1.0,\lambda_3=0.01$. For R1 regularization \cite{mescheder2018training}, we set the parameter $\gamma=10$ at the beginning, and decreasing to $\gamma=5$ gradually.

In conditional experiments, we train our model with a batch size of 16, with learning rate 2.5e-3 and 2e-3 for the generator and discriminator, respectively. We use two-stage training strategy, \ie, training 3.6M images at the resolution of $64^2$ and additional 2.4M images at the resolution of $128^2$.

\subsection{Dataset Details}

\noindent{\bf FFHQ.} FFHQ \cite{karras2019style} is a human-face dataset containing 70K images. In our experiments, we use the preprocessed version provided by EG3D \cite{chan2022efficient}, containing 69957 images with horizontal augmentation.

\noindent{\bf AFHQv2-Cat.} AFHQv2 \cite{choi2020stargan} contains images of animal faces including cats, dogs and wildlife. In our experiments, we only use the set of cats with approximately 5K images, and use the preprocessing method provided by EG3D \cite{chan2022efficient}.
\begin{figure*}
    \centering
    \vspace{-2ex}
    \includegraphics[width=0.85\textwidth,height=220mm]{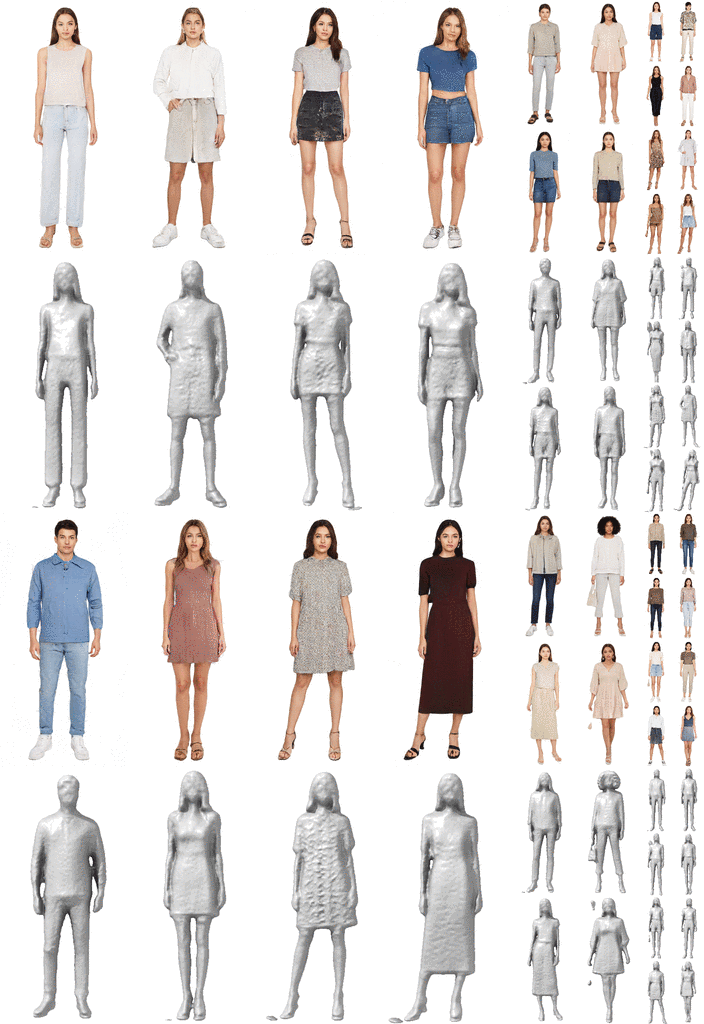}
    \vspace{-1ex}
    \captionof{figure}{\textbf{Images and geometry for seeds 0-31 on SHHQ \cite{fu2022stylegan} $512^2$.} Sampled with truncation \cite{karras2019style} $\psi=0.5$. We crop the image to the resolution of $512\times256$ for better visualization (the background is always white).}
    \vspace{-1ex}
    
\label{fig:shhq_uncond}
\end{figure*}
\noindent{\bf SHHQ.} SHHQ \cite{fu2022stylegan} contains images of full-body humans. The total number of SHHQ is approximately 220K, we use complete dataset to train our models. However, considering that there might be some additional advantages provided by the data volume, we do extra experiments on 110K and 40K images, qualitative results are shown in Fig. \ref{fig:abdatavol}. For conditional task, we use an off-the-shelf human pose estimator PARE \cite{kocabas2021pare} to get human-pose parameters and camera parameters.

\subsection{Visualization of Geometry}

The visualized shapes shown in the manuscript are all generated by UCSF ChimeraX \cite{goddard2018ucsf}.

\section{Additional experiments}
\label{ade}

\begin{figure}[t]
   \centering
   \setlength{\itemsep}{0pt}
\setlength{\parsep}{0pt}
   \includegraphics[width=1.0\linewidth]{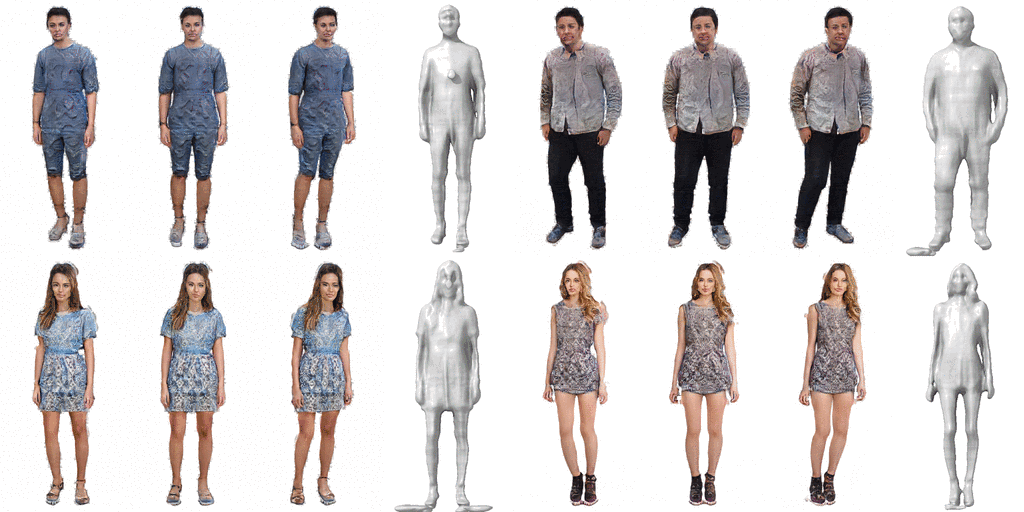}
   \caption{\textbf{Rendered images and geometry with model trained on 40K SHHQ \cite{fu2022stylegan}.} Trained on the dataset with less volume, our model also synthesizes high-quality images with fine geometry.}
\label{fig:abdatavol}
\end{figure}


\noindent{\bf Ablation study on data volume.} As depicted before, we do majority of our experiments and baselines that with official training codes on the complete SHHQ \cite{fu2022stylegan}. However, as an representation, the  effectiveness of orthoplanes should not be related to the data volume. As shown in Fig. \ref{fig:abdatavol}, training on a small subset SHHQ, our approach can obtain high-quality images and geometry as well. For 40K SHHQ, we train our model for 2.8M images rather than 5M images. 

\begin{table}
\tiny
\begin{center}
\resizebox{\linewidth}{!}{
\begin{tabular}{c c c c}
\hline
 & \tiny{\#.Image} & \tiny{FID~$\downarrow$} \\
\hline
{StyleSDF \cite{or2022stylesdf}}  &{220K} &{33.29}\\
{StyleNeRF \cite{gu2021stylenerf}}  &{220K} &{7.60}\\
{EG3D \cite{chan2022efficient}}  &{220K} &{5.79}\\
{Ours}  &{220K} &{4.18}\\
{EG3D \cite{chan2022efficient}}  &{110K} &{14.1}\\
{Ours}  &{110K} &{10.9}\\
{Ours}  &{40K} &{11.2}\\
\hline
\end{tabular}}
\end{center}
\caption{\textbf{Quantitative evaluation for different methods trained on SHHQ with different data volumes.} FID are computed between 50K synthesized images and all visible real images.}
\label{uf}
\vspace*{-6mm}
\end{table}

\begin{figure}[t]
   \centering
   \setlength{\itemsep}{0pt}
\setlength{\parsep}{0pt}
   \includegraphics[width=1.0\linewidth]{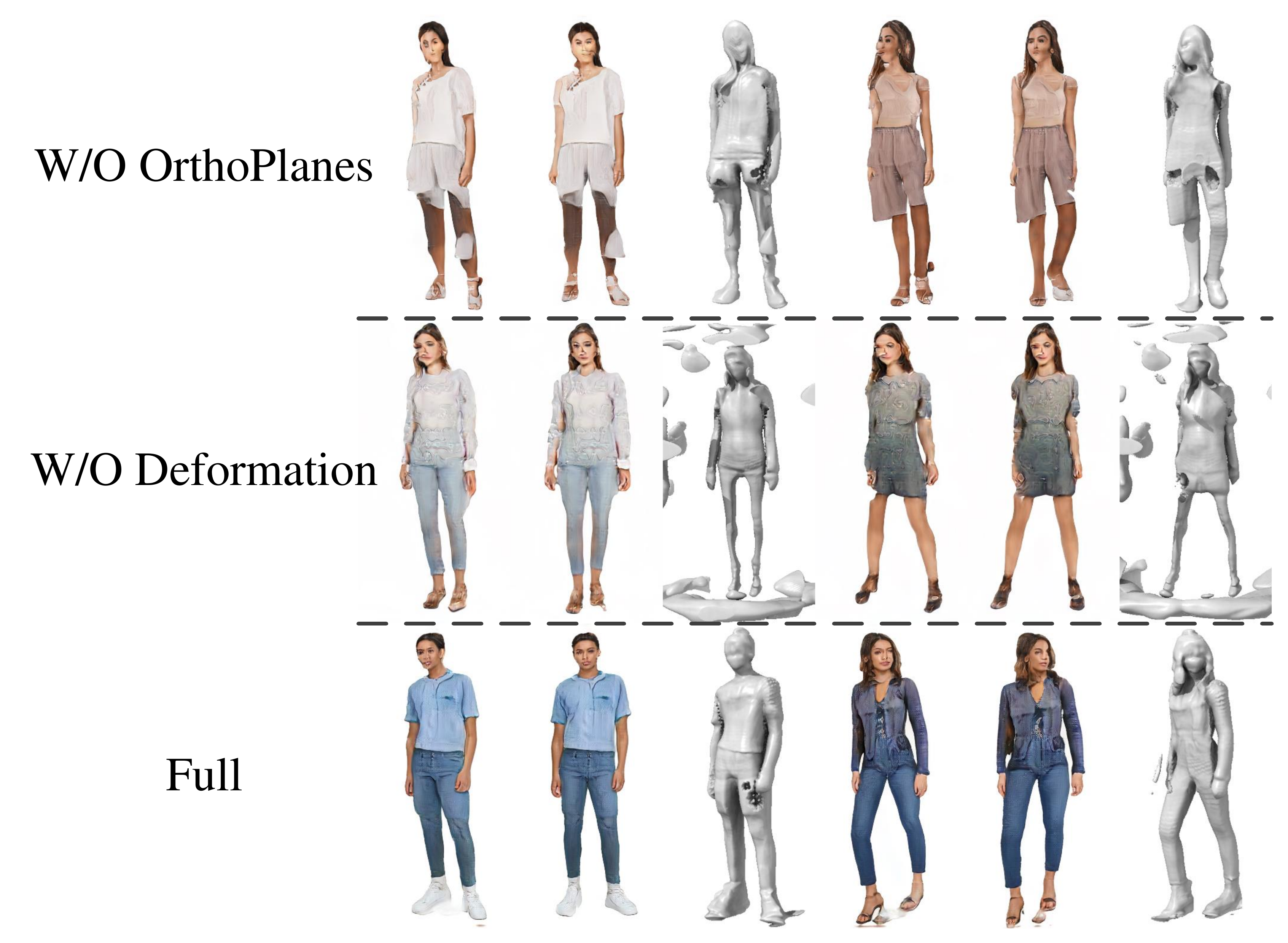}
   \caption{\textbf{Ablation study on pose-conditional human image synthesis.} It's obvious that the model with both of modules generates highest-quality multi-view-consistent images with best geometry.}
\label{fig:abcond}
\end{figure}

\noindent{\bf Ablation study on orthoplanes representation.} In conditional tasks, we train a model with tri-plane as canonical-space codebook generator. As shown in Fig. \ref{fig:abcond}, with less explicit information, it's obvious that geometry generated by tri-plane-based generator lack of details. Furthermore, as the view rolls, both of the appearance and pose are inconsistent, especially the region of face.

\noindent{\bf Ablation study on point-wise deformation.} In our conditional tasks, the deformation module is designed to tackle two problems. First, considering that there might be collision between warped points, which may cause the instability in the training process. Based on point-wise deformation, each point is with its unique offset, mitigating potential conflicts. Second, for clothed human-bodies, there are some non-rigid elements like cloth wrinkles caused by pose, for which the influence of pose should be considered during warping points to the canonical space. As shown in Fig. \ref{fig:abcond}, it's obvious that there are lots of artifacts without deformation module, representing occurred problem of incorrect mapping with inverse LBS only.

    

\section{Additional visual results}
\label{avr}

\subsection{Unconditional image synthesis}

\noindent {\bf Uncurated examples synthesized with FFHQ.} Fig. \ref{fig:ffhq} provides uncurated examples of human-faces generated by our model, trained from scratch with FFHQ \cite{karras2019style}. Truncation \cite{karras2019style} with $\psi$=0.5 is applied.

\noindent {\bf Uncurated examples synthesized with AFHQv2-Cat.} Fig. \ref{fig:afhq} provides uncurated examples of cat-faces generated by our model, trained from scratch with AFHQv2-Cat \cite{choi2020stargan}. Truncation \cite{karras2019style} with $\psi$=0.5 is applied.

\noindent {\bf Uncurated examples synthesized with SHHQ.} Fig. \ref{fig:shhq_uncond} provides uncurated examples of human bodies generated by our model, trained from scratch with SHHQ \cite{fu2022stylegan}. Truncation \cite{karras2019style} with $\psi$=0.5 is applied.

\begin{figure*}
\centering
\subfigure[StyleSDF \cite{or2022stylesdf}]{
\includegraphics[width=0.75\textwidth, height=35mm]{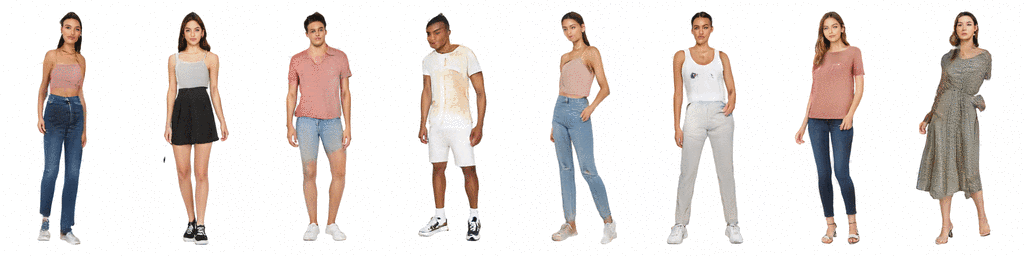} 
}

\subfigure[StyleNeRF \cite{gu2021stylenerf}]{
\includegraphics[width=0.75\textwidth, height=35mm]{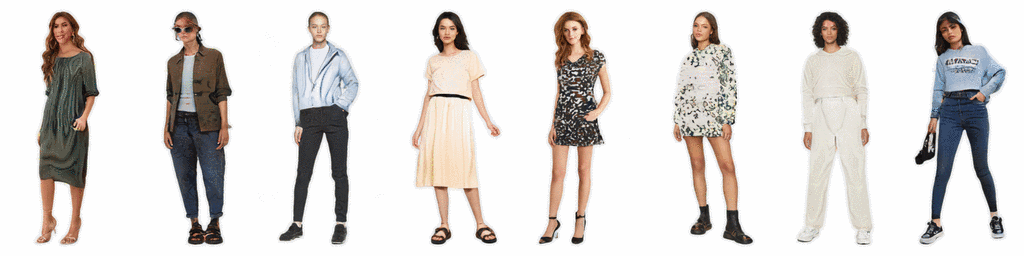} 
}

\subfigure[EG3D \cite{chan2022efficient}]{

\includegraphics[width=0.75\textwidth, height=35mm]{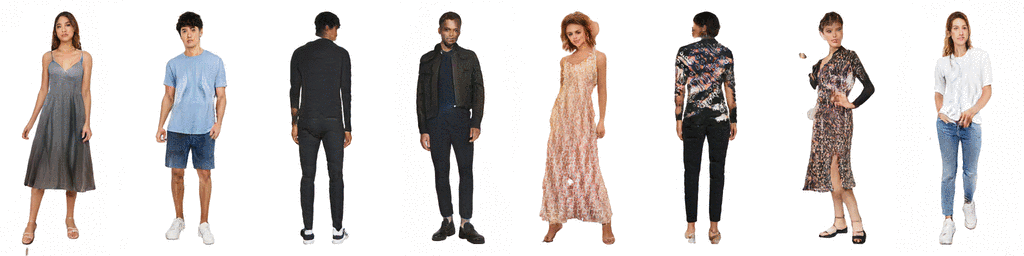} 
}

\subfigure[Ours]{

\includegraphics[width=0.75\textwidth, height=35mm]{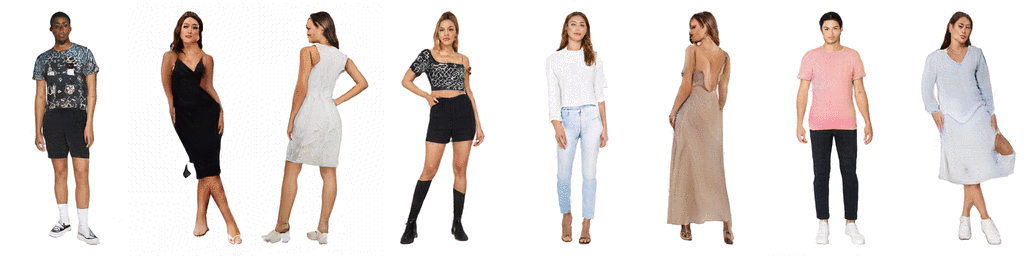} 
}

\caption{\textbf{Qualitative comparison on SHHQ \cite{fu2022stylegan} $512^2$ (Unconditional). For better visualization, we crop all images to the resolution of $512\times256$}. All images are generated without truncation, and all models are trained from scratch.}

\label{fig:com_uncond_shhq}
\end{figure*}

\noindent {\bf Comparison with baselines on SHHQ.} Fig. \ref{fig:com_uncond_shhq} provides

\noindent randomly sampled images produced by StyleNeRF \footnote{\url{https://github.com/facebookresearch/StyleNeRF}} \cite{gu2021stylenerf}, StyleSDF \footnote{\url{https://github.com/royorel/StyleSDF}} \cite{or2022stylesdf}, EG3D \footnote{\url{https://github.com/NVlabs/eg3d}} \cite{chan2022efficient} and our method, respectively. 

\noindent All methods are trained based on their official Pytorch implementations.

\begin{table}
\large
\begin{center}
\resizebox{\linewidth}{!}{
\begin{tabular}{c c c c c c c}
\hline
  &{OP} &{DF} &FID~$\downarrow$ &KID~$\downarrow$ &Depth~$\downarrow$ &PCK~$\uparrow$\\
\hline
& &\checkmark &14.98 &7.71 &0.078 &98.89 \\
&\checkmark & &27.70 &19.3 &0.083 &97.29 \\
&\checkmark &\checkmark &10.36 &5.25 &0.080 &99.25 \\
\hline
\end{tabular}}
\end{center}

\caption{\textbf{Ablation study on pose-conditional human image synthesis.} OP denotes orthoplanes representation proposed in the work, DF denotes the deformation module. All methods are trained for 3.5M images are the neural rendering resolution of $64^2$.}
\vspace*{-6mm}
\end{table}

\subsection{Conditional image synthesis}

\begin{figure*}
    \centering
    \vspace{-2ex}
    \includegraphics[width=0.8\textwidth,height=170mm]{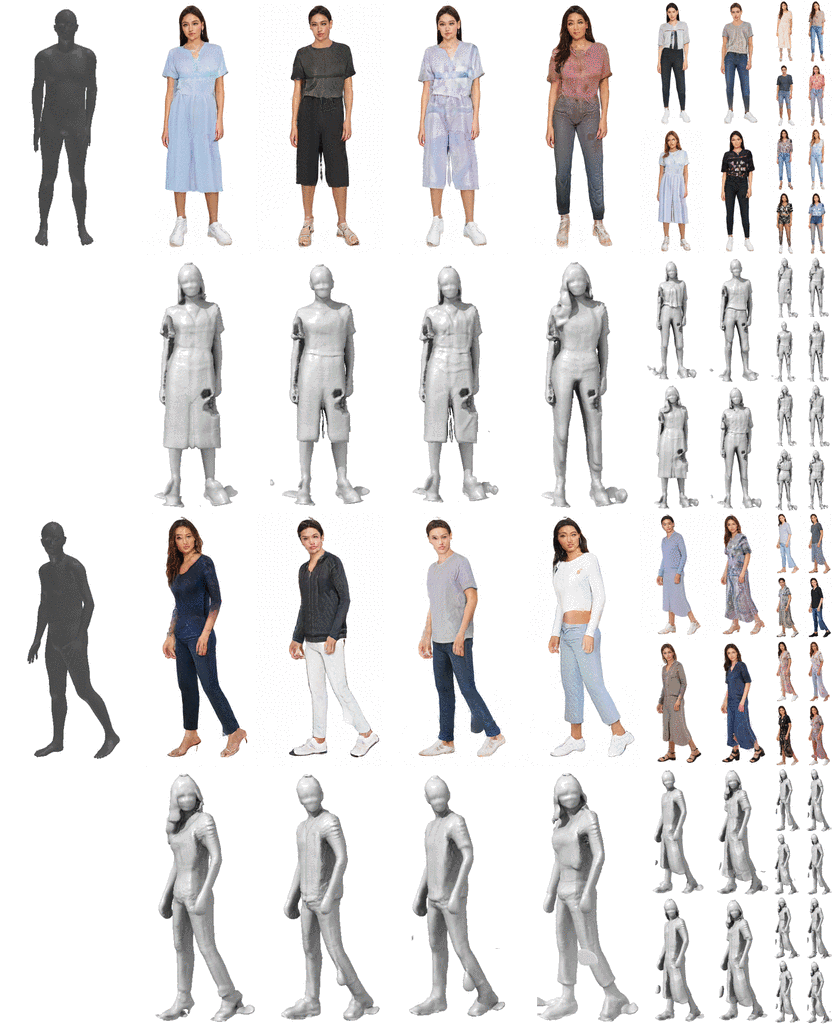}
    \vspace{-1ex}
    \captionof{figure}{\textbf{Images and geometry for seeds 0-31 on SHHQ \cite{fu2022stylegan} $512^2$.} Sampled with truncation \cite{karras2019style} $\psi=0.5$. We crop the image to the resolution of $512\times256$ for better visualization (the background is always white).}
    \vspace{-1ex}
\label{fig:uncurated_pose_shhq}
\end{figure*}

\noindent {\bf Uncurated examples synthesized with SHHQ.} Fig. \ref{fig:uncurated_pose_shhq} provides uncurated examples of pose-conditional human bodies generated by our conditional generator, trained from scratch with SHHQ \cite{fu2022stylegan}. Truncation \cite{karras2019style} with $\psi$=0.5 is applied. The left side shows the mesh with given pose parameters defined by SMPL \cite{loper2015smpl}.

\begin{figure*}
\centering
\subfigure[ENARF-GAN \cite{noguchi2022unsupervised} $128^2$]{
\includegraphics[width=0.75\textwidth, height=60mm]{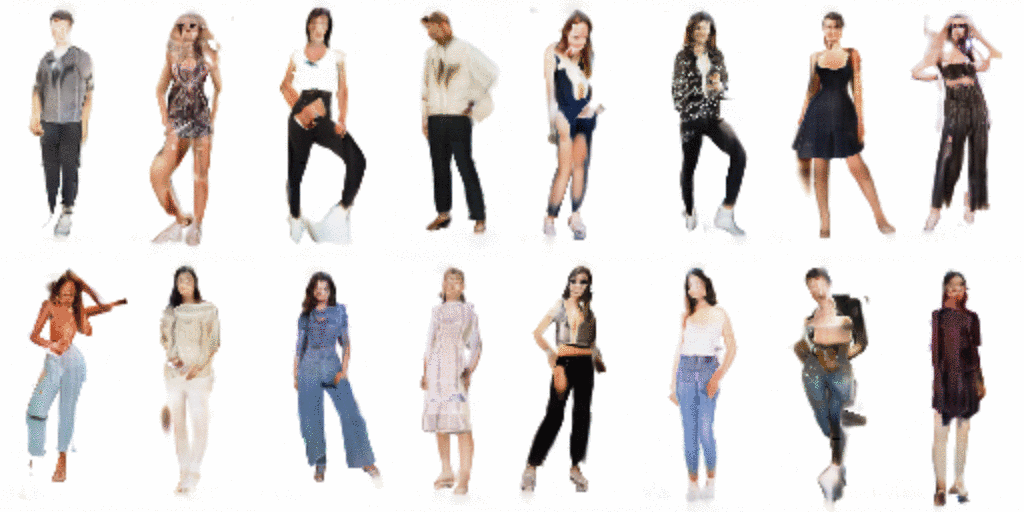} 
}

\subfigure[EVA3D  \cite{hong2022eva3d} $512\times256$]{
\includegraphics[width=0.75\textwidth, height=60mm]{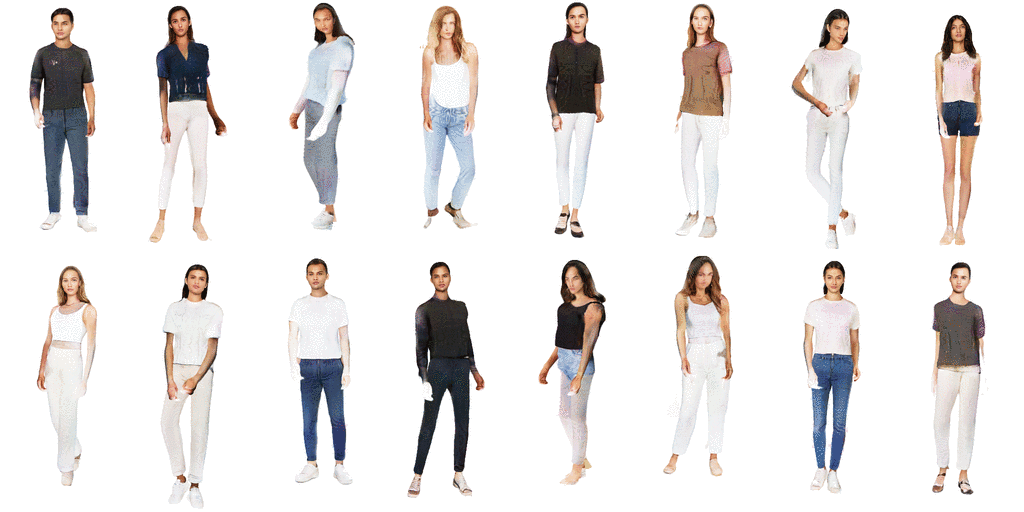} 
}

\subfigure[Ours $512^2$]{

\includegraphics[width=0.75\textwidth, height=60mm]{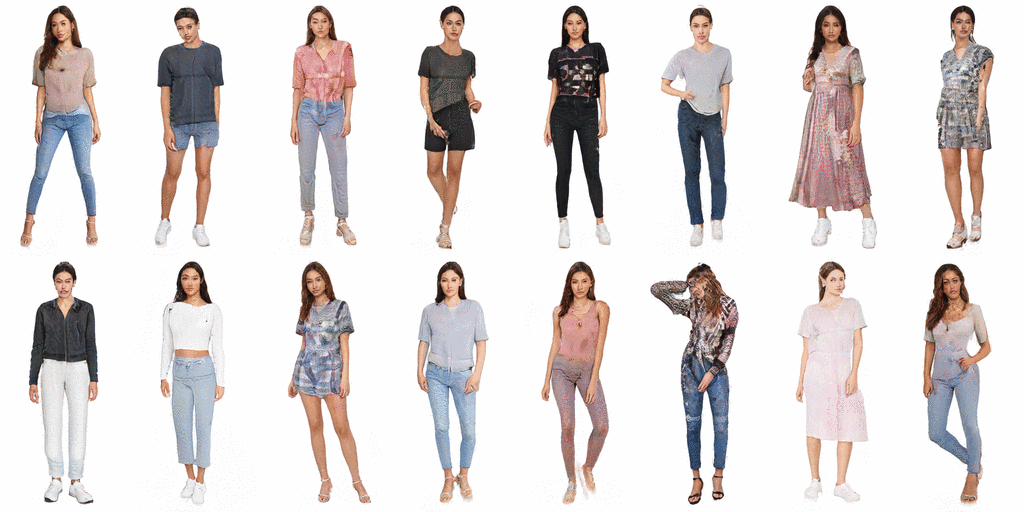} 
}
\caption{\textbf{Qualitative comparison on SHHQ \cite{fu2022stylegan} (Conditional).} For better visualization, we crop the images to the resolution with $H=2W$. Our method generates high-quality results with least artifacts.}
\label{fig:com_pose_shhq}
\end{figure*}

\noindent {\bf Comparison with baselines.} Fig. \ref{fig:com_pose_shhq} provides uncurated RGB results of human bodies controlled by diverse poses, produced by ENARF-GAN \cite{noguchi2022unsupervised}, EVA3D \cite{hong2022eva3d} and our conditional generator, respectively. With composite representation, geometry of EVA3D looks not such natural. Furthermore, the rendered images of the same identity are obviously inconsistent while view changes because of the point-filter used in the model. Our method, based on spatially continuous 3D representation, can generate better geometry. 




\clearpage

\clearpage


\end{document}